\pdfoutput=1

\documentclass[11pt]{article}

\usepackage[preprint]{acl}

\usepackage{times}
\usepackage{latexsym}

\usepackage[T1]{fontenc}

\usepackage[utf8]{inputenc}

\usepackage{microtype}

\usepackage{inconsolata}

\usepackage{graphicx}

\usepackage{multirow}
\usepackage{dirtytalk}
\usepackage{float}
\usepackage{subcaption}

\title{Investigating User Perspectives on Differentially Private Text Privatization}

\author{Stephen Meisenbacher, Alexandra Klymenko, Alexander Karpp, \and Florian Matthes \\
Technical University of Munich\\
School of Computation, Information and Technology \\
Department of Computer Science\\
Garching, Germany\\
\texttt{\{stephen.meisenbacher,alexandra.klymenko,a.karpp,matthes\}@tum.de} \\
}

\begin{document}
\maketitle
\begin{abstract}
Recent literature has seen a considerable uptick in \textit{Differentially Private Natural Language Processing} (DP NLP). This includes DP text privatization, where potentially sensitive input texts are transformed under DP to achieve privatized output texts that ideally mask sensitive information \textit{and} maintain original semantics. Despite continued work to address the open challenges in DP text privatization, there remains a scarcity of work addressing user perceptions of this technology, a crucial aspect which serves as the final barrier to practical adoption. In this work, we conduct a survey study with 721 laypersons around the globe, investigating how the factors of \textit{scenario}, \textit{data sensitivity}, \textit{mechanism type}, and \textit{reason for data collection} impact user preferences for text privatization. We learn that while all these factors play a role in influencing privacy decisions, users are highly sensitive to the utility and coherence of the private output texts. Our findings highlight the socio-technical factors that must be considered in the study of DP NLP, opening the door to further user-based investigations going forward.
\end{abstract}

\section{Introduction}
The pursuit of text privatization under the framework of Differential Privacy (DP) presents a promising, yet challenging task for researchers, who must balance the strong protections DP offers with the ability to retain meaningful utility from textual data \cite{klymenko-etal-2022-differential}. In recent years, numerous works at the intersection of data privacy and Natural Language Processing, better known as privacy-preserving NLP or PPNLP, have tackled this challenge in various methods and techniques leveraging DP for text privatization \cite{hu-etal-2024-differentially}. These range from word replacement methods \cite{10.1145/3336191.3371856}, more advanced autoencoder-based methods \cite{igamberdiev-habernal-2023-dp}, and recent works leveraging LLMs for privatization \cite{utpala-etal-2023-locally}.

Addressing the technical challenges in realizing effective DP text privatization mechanisms has been at the forefront of researchers' goals in the recent literature. Often, researchers proposing new methods must not only prove that a mechanism satisfies DP, but they must also empirically demonstrate that the mechanism can provide some tangible privacy benefit while also producing private texts that are useful and coherent \cite{mattern-etal-2022-limits}. Furthermore, operating in the domain of natural language introduces the complexities of syntactic hierarchy \cite{vu-etal-2024-granularity} and meaningful privacy budgets \cite{igamberdiev-habernal-2023-dp}, as well as clearly delineating the advantages of DP over traditional anonymization \cite{meisenbacher-matthes-2024-thinking} and maintaining reproducibility, explainability, and comparability \cite{igamberdiev-etal-2022-dp, meisenbacher2024comparative}. 

Beyond these complexities, an under-explored aspect of DP in NLP remains measuring human perceptions of DP text privatization. Very few works have extended past the research sphere to engage everyday users in investigating their perspective on what effective DP text privatization actually means. A recent work by \citet{weiss-etal-2024-share} opens the doors to this aspect, taking a \textit{risk-based} approach in quantifying at which privacy budgets (or, the $\varepsilon$ parameter) laypersons are comfortable in sharing their personal text data. Here, it is shown that users are influenced by the perceived risk of misuse of their data, as they are less likely to consent to sharing with higher stated risks.

Despite its important role in leading off the study of human perceptions of DP NLP, we see a number of limitations in the work proposed by \citeauthor{weiss-etal-2024-share}. Firstly, the \textit{risk perception} approach taken by this work is useful in simplifying data sharing scenarios to laypersons, yet it makes no direct connection to actual outputs of privatization mechanisms, thus largely ignoring the crucial factor of \textit{language} in text privatization. Relatedly, the work only considers the \textit{global} DP setup, which distances itself from tangible privatization outputs and abstracts the DP notion away from local users. Because of this, we gain little insight into user opinions and preferences of \textit{local} privatization mechanisms, which comprise a large portion of the recent literature.

We build upon the previous research of \citeauthor{weiss-etal-2024-share} by focusing on these limitations, conducting a user study to investigate perceptions of text-to-text privatization in various data sharing scenarios. We frame our user study in the form of \textit{vignettes}, allowing for richer scenarios in which the users are placed. In these vignettes, we explore the influence of several important factors in local DP text privatization, including mechanism type, privacy budget, sensitivity of scenario, and reason for data collection. The choice of tested mechanisms is guided by a literature review of recent DP NLP works.

Our survey with 721 users from around the world yields interesting insights and perspectives on DP text privatization. Above all, we find that the choice of privatization mechanism \textit{does} matter, and users generally perceive mechanisms producing more coherent and natural outputs as preferable. If outputs are not so, users tend to choose \textit{less} privacy in preference of \say{utility}. Finally, we find that \textit{sensitivity of scenario} and \textit{reason for data collection} are important, but not of primary concern. 

These findings provide a clear call to action for DP NLP researchers, namely to continue to study the perceptions of users, in order to align DP NLP research with real-world perspectives and needs. In this light, we make the following contributions:
\begin{enumerate}
    \itemsep -0.1em
    \item We build upon previous work by investigating user perceptions of DP text privatization.
    \item We are the first to employ a \textit{vignette}-based user study in the context of text privatization.
    \item We share the findings of our study, including statistically significant results leading to recommendations for future DP NLP research.
\end{enumerate}

\section{Related Work}
Several recent works in DP NLP, although focusing on the technical aspect of the topic, point to the need for deeper consideration of the practical implications of DP text privatization. The work of \citet{mattern-etal-2022-limits} critiques earlier word-level DP mechanisms, uncovering the issues of grammatical correctness and semantic coherence, a challenge more recent works have addressed \cite{10.1145/3485447.3512232,utpala-etal-2023-locally,10.1145/3664476.3669926}. Specifically considering syntactics, \citet{vu-etal-2024-granularity} demonstrate the importance of granularity, or syntactic hierarchy, especially in real-world data sharing scenarios. Quantifying and addressing these challenges becomes important to demonstrating the practical applicability of DP NLP \cite{meisenbacher-matthes-2024-thinking}, especially in light of more real-world challenges such as explainability and transparency \cite{klymenko-etal-2022-differential, igamberdiev-habernal-2023-dp, igamberdiev-etal-2024-dp}, as well as reproducibility and comparability \cite{igamberdiev-etal-2022-dp, meisenbacher2024comparative}.

Particularly investigating the human aspect of text privatization, little work outside of \citet{weiss-etal-2024-share} has been performed. However, beyond the field of NLP, usable privacy research has been considerably more active in exploring user perspectives on DP. Several works explore which communication methods are most effective in explaining DP to end users \cite{10.1145/3460120.3485252,10.1145/3548606.3560693,10.5555/3620237.3620328}, which generally find that how DP is explained to users is important in fostering their understanding of the risks and implications. \citet{281270} find that high-level abstractions of DP may lead to misunderstandings or false expectations about DP, and \citet{10.1145/3555762} conclude that sometimes explanations have little effect on users' willingness to share data. Interestingly, one work \cite{xiong2020towards} shows that local DP (LDP) concepts are more understandable than DP, and that in the LDP case, users exhibit more willingness to share data. 

We are motivated by these previous works in the DP field, particularly to provide more clarity on user perceptions of DP NLP methods. In light of the importance found by these previous works on the \textit{method} of investigating user perspectives, we focus in this work on showing direct \textit{outputs} of LDP text privatization mechanisms to users, in the form of understandable \textit{vignettes}, as employed by \citet{10.5555/3620237.3620328} for DP. With these, we are able to analyze different factors in the context of DP text privatization, particularly those affecting user perceptions of text privatization outputs.

\section{Experimental Design}
We outline our experimental design, which consists of an initial literature review, followed by survey implementation, and finally, the survey conduction.

\begin{table*}[ht]
\centering
\small
\begin{tabular}{c|c|c|p{7cm}}
\hline
\textbf{Mechanism Type} & \textbf{Syntactic Level} & \textbf{DP Definition} & \multicolumn{1}{c}{\textbf{Sources}} \\ \hline
\multirow{4}{*}{\textbf{\begin{tabular}[c]{@{}c@{}}Word-level\\Noise Addition\end{tabular}}} & \multirow{4}{*}{Word} & DP & \cite{weggenmann2018syntf} \\ \cline{3-4} 
 &  & MDP & \cite{fernandes2019generalised, xu-etal-2020-differentially} \\ \cline{3-4} 
 &  & LDP & \cite{bollegala-etal-2023-neighbourhood} \\ \cline{3-4} 
 &  & MLDP & \cite{8970912, 10.1145/3336191.3371856, lyu-etal-2020-differentially, xu2021density, xu-etal-2021-utilitarian,Imola2022,arnold-etal-2023-driving,arnold-etal-2023-guiding,carvalho2023tem} \\ \hline
\multirow{2}{*}{\textbf{Binary Embeddings}} & \multirow{2}{*}{Word} & LDP & \cite{10.1145/3397271.3401260} \\ \cline{3-4} 
 &  & MLDP & \cite{carvalho2021brrpreservingprivacytext} \\ \hline
\multirow{3}{*}{\textbf{\begin{tabular}[c]{@{}c@{}}Exponential\\Mechanism-\\Based\end{tabular}}} & Token & LDP & \cite{Chen2022ACT, meisenbacher-etal-2024-dp} \\ \cline{2-4} 
 & Word & UMLDP & \cite{yue} \\ \cline{2-4} 
 & Sentence & DP & \cite{meehan-etal-2022-sentence} \\ \hline
\multirow{7}{*}{\textbf{\begin{tabular}[c]{@{}c@{}}Autoencoder-Based\\(AE)\end{tabular}}} & \multirow{2}{*}{Word} & LDP & \cite{habernal-2021-differential, plant-etal-2021-cape, krishna-etal-2021-adept, maheshwari-etal-2022-fair} \\ \cline{3-4} 
 &  & MLDP & \cite{feyisetan-kasiviswanathan-2021-private} \\ \cline{2-4} 
 & \multirow{2}{*}{Sentence} & $(\varepsilon,\delta)$-DP & \cite{bo-etal-2021-er} \\ \cline{3-4} 
 &  & MLDP & \cite{10.1145/3543507.3583512} \\ \cline{2-4} 
 & \multirow{3}{*}{Document} & \begin{tabular}[c]{@{}c@{}}LRDP, \\ $(\varepsilon,\delta)$-DP\end{tabular} & \cite{10.1145/3485447.3512232} \\ \cline{3-4} 
 &  & DP & \cite{10.1145/3342220.3344925} \\ \cline{3-4} 
 &  & LDP & \cite{igamberdiev-habernal-2023-dp} \\ \hline
\multirow{2}{*}{\textbf{LLM-Based}} & \multirow{2}{*}{Token} & \multirow{2}{*}{LDP} & \multirow{2}{*}{\cite{mattern-etal-2022-differentially, utpala-etal-2023-locally}} \\
 &  &  & 
\end{tabular}
\caption{A selection of DP text privatization methods, resulting from our scoping literature review.}
\label{tab:methods}
\end{table*}

\subsection{Literature Review}
As our survey study is focused on presenting users with tangible outputs from DP text privatization mechanisms, our first step included an unstructured scoping literature review \cite{munn2018systematic}, with the goal of identifying available DP text privatization methods for inclusion in our survey. This review was largely aided by a recent survey \cite{hu-etal-2024-differentially}, which we augmented with DP NLP papers published after this work. In particular, we excluded the methods denoted by \citeauthor{hu-etal-2024-differentially} as \say{Gradient Perturbation} methods, as well as those that involve DP vector perturbation in training or fine-tuning. In this way, we only include methods that result in private texts as a direct result of DP.

The results of our review are presented in Table \ref{tab:methods}, which delineates methods into five distinct mechanism types, the linguistic level on which the mechanism operates, and its DP notion. For the purposes of this work, we choose four representative methods, excluding the category of \textit{Binary Embeddings} due to its inability to produce natural language outputs. The selection of the following four methods was performed to (1) represent a diversity in syntactic level, (2) focus solely on LDP, and (3) prioritize newer works:

\begin{itemize}
    \itemsep -0.1em
    \item \textbf{Truncated Exponential Mechanism (\textsc{TEM})} \cite{carvalho2023tem}: word-level Metric LDP mechanism.
    \item \textbf{\textsc{DP-MLM}} \cite{meisenbacher-etal-2024-dp}: token-level LDP mechanism leveraging masked language models.
    \item \textbf{\textsc{DP-Prompt}} \cite{utpala-etal-2023-locally}: token-level LDP leveraging LLMs for paraphrasing.
    \item \textbf{\textsc{DP-BART}} \cite{igamberdiev-habernal-2023-dp}: document-level LDP mechanism leveraging the BART model \cite{lewis-etal-2020-bart}.
\end{itemize}

\begin{figure*}
    \centering
    \includegraphics[scale=0.6]{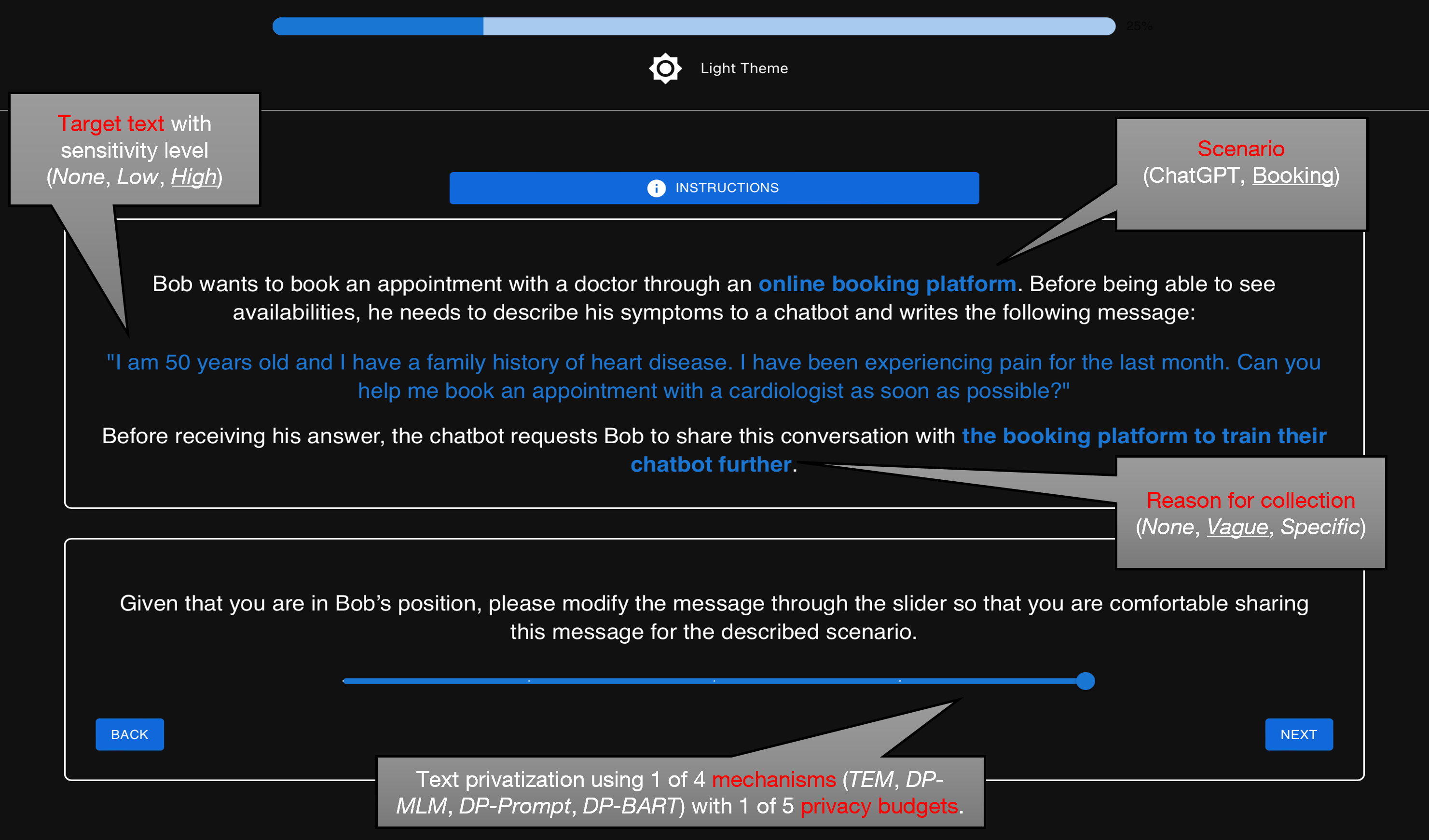}
    \caption{An example of a \textit{vignette} on our survey platform. The annotations in the figure indicate the different \textit{factors} of our FSM model, where underlined \textit{treatments} are those depicted in the example. Participants were presented first with the original ($\varepsilon = \infty$) text, and then could proceed to use the slider to consider privatized counterparts.}
    \label{fig:vignette}
\end{figure*}

\subsection{Survey Design}
In order to learn about user perspectives on DP text privatization, we designed a survey study to answer the following research question:

\begin{quote}
    \textit{What insights can be gained about the factors influencing user perception of \ differentially private text privatization?}
\end{quote}

As previously mentioned, we chose to design our survey in the form of \textit{vignettes} \cite{atzmuller2010experimental}. Vignettes use short descriptions to prompt respondents to place themselves in a scenario and to respond accordingly. These scenarios allow for more realistic contexts, and they can be particularly effective for exploring sensitive topics \cite{auspurg2014factorial}, such as privacy. Specifically, we employ the Factorial Survey Method (FSM), in which multiple factors, or dimensions, are varied and tested, allowing for an analysis of the causal relationship of these factors in influencing user responses \cite{auspurg2014factorial}.

\begin{table}[t!]
\centering
\small
\resizebox{\linewidth}{!}{
\begin{tabular}{c|p{0.3\linewidth}|p{0.60\linewidth}}
\hline
\textbf{Factor} & \textbf{Level} & \textbf{Treatment} \\ \hline
 & Word & \textsc{TEM} \\
Mechanism & Token & \textsc{DP-MLM} \\  & Document, AE & \textsc{DP-BART} \\
 & Document, LLM & \textsc{DP-Prompt} \\ \hline
 & \textsc{TEM} & $\varepsilon \in \{1.6, 2.4, 2.8, 3, \infty\}$ \\
Privacy & \textsc{DP-MLM} & $\varepsilon \in \{20, 35, 50, 125, \infty\}$ \\
Budget ($\varepsilon$) & \textsc{DP-Prompt} & $\varepsilon \in \{35, 45, 50, 65, \infty\}$ \\
 & \textsc{DP-BART} & $\varepsilon \in \{300, 400, 700, 1400, \infty\}$ \\ \hline
Reason for & None & No reason given \\
data & Model Training & Vague reason given \\
collection  & Privacy Protection & Specific reason given \\ \hline
 & None & 0 personal attributes \\
Sensitivity & Low & 1 personal attribute \\
of text & High & 2 personal and 1 sensitive attribute \\ 
\end{tabular}
}
\caption{An overview of our FSM model's factors, levels, and treatments.}
\label{tab:fsm}
\end{table}

Our FSM model consists of four factors: \textit{mechanism}, \textit{privacy budget}, \textit{reason for data collection}, and \textit{sensitivity of data}. In an FSM study, each factor contains a number of \textit{levels}, which are realized in the survey by \textit{treatments}. The factors, levels, and treatments are summarized in Table \ref{tab:fsm}.

\paragraph{Vignette Creation.}
We sought to create vignettes that are both understandable and relatable to users, as well as representative of some plausibly sensitive data sharing scenario. The first step involved our research team brainstorming such scenarios, which resulted in eight distinct candidates. Each vignette was created with a similar structure: (1) introduction to the vignette, (2) presentation of the target text to be privatized, and (3) reason for data collection, explained below. In drafting the vignettes, we followed the best practices of \citet{EVANS2015160}, namely to be clear and concise, use present tense, and keep a consistent structure across vignettes.

According to our FSM model, we then proceeded to draft different versions of the eight vignette scenarios, focusing on the two factors of \textit{reason for data collection} and \textit{sensitivity of data}. For the former, we included a text at the end of the vignette informing a user for which purpose the data was to be shared (see Table \ref{tab:fsm}). To vary data sensitivity, we modified the text to be shared (i.e., privatized) with personal or sensitive attributes, as defined in Recitals 51 to 56 of the GDPR. Concretely, the \textit{None} treatment contained no personal attributes, the \textit{Low} treatment contained one personal attribute, and the \textit{High} treatment contained two personal attributes and one sensitive attribute (e.g., medical condition). All eight vignette candidates can be found in Appendix \ref{sec:vote}.

Our goal was to narrow the selection down to two scenarios for the survey study, primarily to keep the scope within reason. To accomplish this, we ran a committee vote in the form of a survey. In this survey, we asked respondents to rank each of the scenarios in terms of \textit{relevance}, \textit{plausibility}, and \textit{understandability} for a data sharing scenario. The ranking was performed on a five-point Likert scale (\textit{strongly disagree} to \textit{strongly agree}). The committee consisted of 12 close research colleagues.

The top two scoring vignettes both involved a \textit{health} scenario, one where a user is researching a medical condition with the help of ChatGPT, and the other where the user is interacting with an online booking platform chatbot to book a doctor's appointment. Although both vignettes operate in a similar sensitive domain, we decided to adhere to the committee vote without further adjustments.

With this, our study thus consisted of a vignette domain of 18 vignettes per mechanism (2 scenarios $\times$ 3 sensitivity levels $\times$ 3 collection reasons), resulting in an overall collection of 72 vignettes. 

\paragraph{Mechanisms and Budgets.}
For each of the chosen mechanisms, we selected five privacy budgets ($\varepsilon$), the last of which was $\infty$, i.e., the original text. To ensure comparability between mechanisms, we decided to fix the remaining budgets based on the \textit{average semantic similarity} between original and private text, given a mechanism and budget. We set four target similarities of $\{0.2, 0.4, 0.6, 0.8\}$, and proceeded to define an \textit{$\varepsilon$ range} for each mechanism, given roughly by the minimum and maximum values tested in the original papers, with 30 steps within this range. For each of these 30 values, we ran the mechanism 20 times on our two vignette target texts, and used a \textsc{sentence-transformers/all-MiniLM-L6-v2} \cite{reimers-gurevych-2019-sentence} to compute the average cosine similarity. Then, the closest $\varepsilon$ value to each of our targets was chosen, resulting in the values in Table \ref{tab:fsm}. For the actual survey implementation (discussed next), the closest of the 20 texts to each target value was preserved; thus, the five texts used in each vignette are fixed. The privatized texts for each mechanism are provided in Appendix \ref{sec:private}.

\paragraph{Survey Platform Implementation.}
Due to the unique setup of our survey study, we decided that a custom web application would be best suited for our needs, rather than relying on existing online services. Most important was the facilitation of our \textit{$\varepsilon$ slider} functionality, where survey respondents could dynamically view the privatization of the target texts by switching between the five privacy budget values. For the application, we opted to use React\footnote{\url{https://react.dev/}} for the frontend and Node.js\footnote{\url{https://nodejs.org/}} for the backend. The flow of the survey was as follows:
\begin{enumerate}
    \itemsep -0.1em
    \item \textit{Introduction}: welcome / detailed instructions.
    \item \textit{Demographics}: information about gender, age, country, education, and occupation.
    \item \textit{IUIPC-10}: baseline questions about the respondent's general privacy opinion, using the Internet User Information Privacy Concerns Questionnaire \cite{malhotra2004internet}, as utilized by \citet{weiss-etal-2024-share}.
    \item \textit{Vignettes}: as exemplified in Figure \ref{fig:vignette}. We customized the vignette selection process to ensure that all vignettes were sampled equally.
    \item \textit{Open Feedback}: three free text fields asking for further comments on the survey.
\end{enumerate}

The system architecture diagram of the survey web application can be found in Appendix \ref{sec:webapp}.

\paragraph{Participant Recruitment.}
We ran initial pilot tests with contacts in our personal network (n=41). The goal of these pilots was to estimate the total time of completion, as well as to identify and correct any ambiguities or technical issues. Before the pilots, we set an initial target goal that each unique vignette (72 in total) would be answered approximately 100 times each, for a total of 7200 responses needed. We set each survey to contain 10 vignettes; thus, our target sample size was 720. 

The pilot tests identified no technical issues; however, improvements were made to the instructions to clarify to participants that text privatization would occur \textit{locally}, and this process would not affect the quality of the response (i.e., from ChatGPT or the booking chatbot), as privatization in our context only affects the data \textit{stored}. We measured an average completion time of around 10 minutes.

For the main study following the pilot tests, we used the Prolific\footnote{\url{https://www.prolific.com/}} platform for recruitment. We did not limit participants by geographic region; however, we did require fluency in English and at least a high school diploma. We set the study for 680 total participants, paid $\pounds 1.50$ for survey completion (rate of $\pounds 9$/hr), marked by Prolific as a \say{good} wage. For the Prolific segment, it is important to note that two \say{attention checks} were inserted into the vignette portion, as required by the platform. Failure of these checks disqualified participants from compensation. In our survey, these checks took the form of normal vignettes, but with the explicit instruction to choose the slider value 3 (i.e., the middle value). One failed attention check was allowed, but two led to disqualification.

\section{Results}
We present the results of our user study, prefaced by our tested hypotheses and augmented by an analysis of our respondents' privacy preferences.

\subsection{Hypotheses}
\label{sec:hypo}
To empirically measure the factors influencing user willingness to share textual data, particularly under local DP privatization, we construct a research model with three primary hypotheses, as follows:
\begin{enumerate}
    \itemsep 0em
    \item[\textbf{H1}:] A higher sensitivity level in texts will result in a lower chosen privacy budget ($\varepsilon$), leading to increased preference of DP privatization.
    \item[\textbf{H2}:] Mechanisms that lack linguistic and/or semantic preservation will lead to an increase in the chosen privacy budget.
    \item[\textbf{H3}:] Providing a reason for data collection will increase the likelihood of users sharing their data under lower privacy regimes (higher $\varepsilon$), compared to not providing a reason.
\end{enumerate}

\begin{figure}[t!]
    \centering
    \includegraphics[scale=0.4]{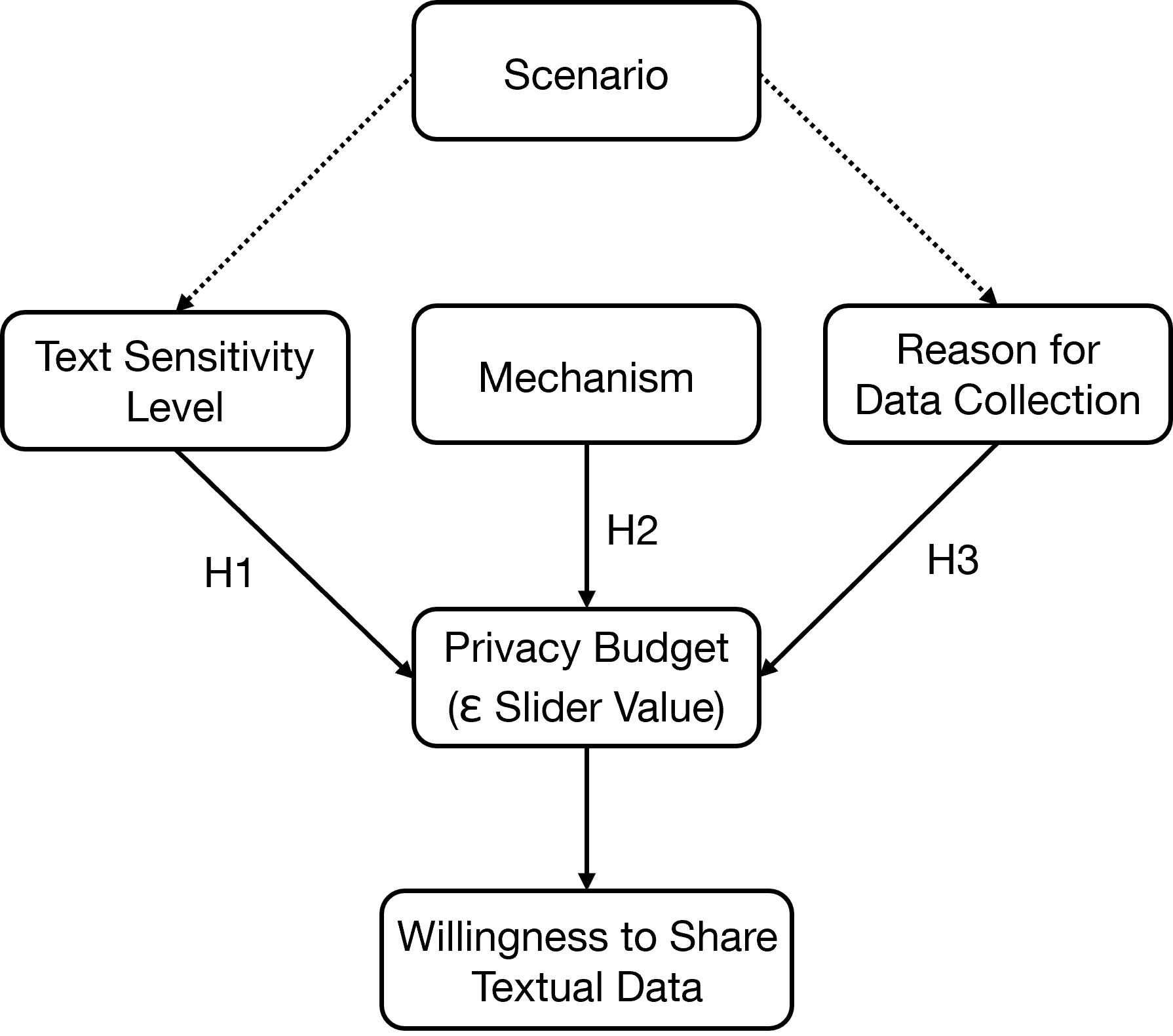}
    \caption{Our research model for the FSM study.}
    \label{fig:model}
\end{figure}

We hypothesize that these factors of \textit{data sensitivity} (H1), \textit{mechanism} (H2), and \textit{data collection reason} (H3) will impact a user's choice of privatization level (governed by $\varepsilon$), thereby influencing a user's willingness to share their textual data. Note that in the context of our work, we consider \textit{generative} methods (i.e., \textsc{DP-Prompt} and \textsc{DP-BART}) to preserve linguistics and semantics, and \textit{non-generative} methods (i.e., \textsc{DP-MLM} and \textsc{TEM}) as lacking the ability to do so. 

The research model is illustrated in Figure \ref{fig:model}.

\subsection{Participant Demographics}
Our conducted survey consisted of 721 total respondents, including friends and family (n=41) and Prolific participants (n=680). Of these, 53.5\% identified as female (n=386), 45.8\% as male (n=330), and five respondents preferred not to answer. The survey participants were uniformly distributed across age ranges, including under 18 (n=2), 18-24 (n=151, 20.9\%), 25-34 (n=321, 44.5\%), 35-54 (n=193, 26.8\%), and over 55 (n=54, 7.5\%).

The survey respondents were located in 41 different countries across six continents. The top-5 most frequent countries were South Africa (n=226, 31.3\%), United Kingdom (n=132, 18.3\%), Italy (n=46, 6.4\%), United States (n=45, 6.2\%), and Germany (n=39, 5.4\%). Overall, the most respondents came from Europe (n=384, 53.3\%), in addition to Africa (n=253, 35.1\%), North America (n=55, 7.6\%), Asia (n=16, 2.2\%), South America (n=9, 1.2\%), and Australia (n=4, 0.6\%).

The largest group of respondents work in the industry (n=300, 41.6\%), and the rest indicated being a student (n=123, 17.1\%), unemployed (n=79, 6.5\%), self-employed (n=50, 6.9\%), in research (n=48, 6.7\%), or \say{Other} (n=121, 16.8\%). Nearly half of the respondents hold a Bachelor's degree (n=354, 49.1\%), while others hold a Master's (n=147, 20.4\%), High School or equivalent degree (n=127, 17.6\%), apprenticeship (n=46, 6.4\%), doctorate (n=27, 3.7\%), or \say{Other} (n=20, 2.8\%).

\begin{table}[t!]
\centering
\small
\resizebox{\linewidth}{!}{
\begin{tabular}{l|c|c|c}
 & \multicolumn{3}{c}{\textbf{$\overline{x}$ ($\sigma$)}} \\ \cline{2-4} 
\textbf{Category} & \textbf{\scriptsize\citet{malhotra2004internet}} &  \textbf{\scriptsize\citet{gross2021validity}} & \textbf{\scriptsize Our study} \\ \hline
Control & 5.67 (1.06) & 5.87 (0.87) & 6.09 (0.85) \\
Awareness & 6.21 (0.87) & 6.39 (0.65) & 6.53 (0.67) \\
Collection & 5.63 (1.09) & 5.50 (1.09) & 5.91 (1.16) \\ \hline
IUIPC-10 & 5.84 (1.01) & 5.93 (0.67) & 6.18 (0.66)
\end{tabular}
}
\caption{Comparison of IUIPC-10 Results from two previous works and our observed sample. Values given represent average Likert scale scores (1-7), with standard deviations provided in parentheses.}
\label{tab:IUIPC}
\end{table}
 
\subsection{IUIPC Results}
In Table \ref{tab:IUIPC}, we present a comparative illustration of the observed IUIPC scores from our survey study, juxtaposed with results from previous works, including the original paper \cite{malhotra2004internet} and a more recent study \cite{gross2021validity}. As can be seen, our study population self-reported as very privacy-conscious, scoring higher in each IUIPC-10 sub-scale than the referenced previous works. In addition to being a relevant basis for the ensuing analysis and discussion, these results imply a growing privacy awareness globally, which can be attributed to increasing attention paid to large-scale data processing, particularly related to modern AI. Most notably, the \textit{Awareness} category received high scores, showing that knowledge of data collection and processing by third parties is a timely subject and is on people's minds. We refer the reader to Appendix \ref{sec:iuipc} for a more in-depth analysis of the IUIPC results.

\begin{figure*}[htbp]
    \centering
    \includegraphics[scale=0.26]{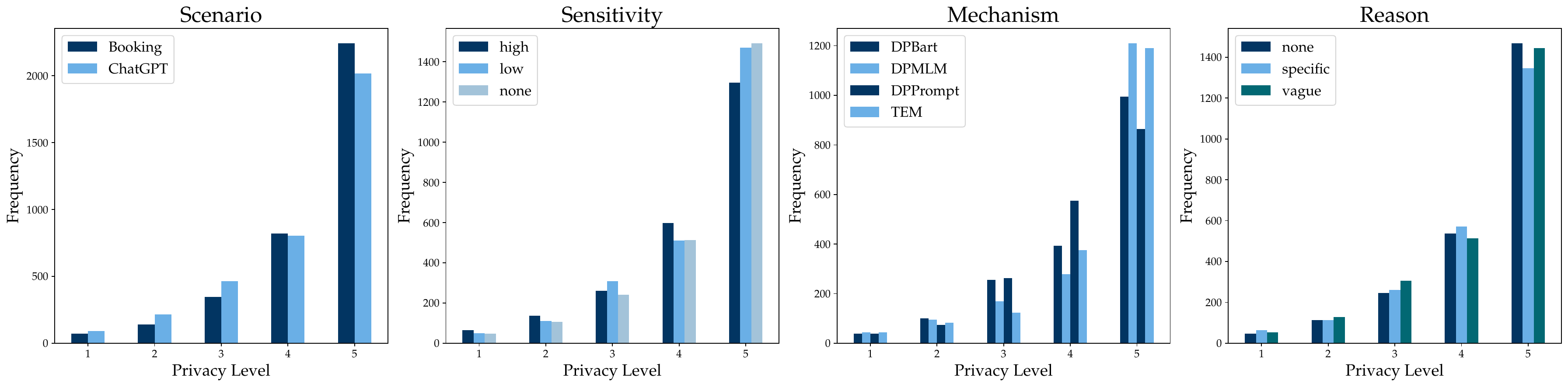}
    \caption{Raw frequency of privacy level responses (1-5) per each tested factor.}
    \label{fig:variables}
\end{figure*}

\subsection{Analysis of Vignette Responses}
The analysis of our survey study responses is centered on the influence of our chosen factors (i.e., those in our hypotheses) on the selection of privacy level for DP text privatization, indicated by the selected slider value in our vignettes. In the following, we perform statistical tests to determine the significance of these factors, as well as to support or refute our hypotheses.

\subsubsection{Initial Testing}
An initial review of the overall vignette responses revealed a very skewed distribution, with higher slider values (i.e., higher $\varepsilon$ values) being chosen far more often. In particular, where slider value 5 corresponds to $\varepsilon = \infty$ and slider value 1 is the lowest chosen budget per mechanism, we observed the following out of 7210 responses: 5 (n=4260, 59.1\%), 4 (n=1622, 22.5\%), 3 (n=810, 11.2\%), 2 (n=354, 4.9\%), and 1 (n=164, 2.3\%). As this clearly does not follow a normal distribution (shown in-depth in Appendix \ref{sec:normality}), we opted to use non-parametric tests for our analysis, i.e., those that do not rely on assumptions about the data distribution.

\subsection{Chi-squared Testing}
We first examined the relationships between the dependent variable, or chosen privacy budget, with our four independent variables: \textit{scenario}, \textit{sensitivity}, \textit{mechanism}, and (data collection) \textit{reason}. Here, the privacy budget represents a categorical variable on a scale of 1 to 5 (slider values).

We chose to conduct Chi-squared tests to determine the influence of our independent variables on the chosen privacy budget value. In addition to our data being non-normally distributed, these tests were reasonable to conduct since they are well-suited to test relationships between \textit{categorical} variables (as our variables are). Furthermore, we compare the observed frequencies of privacy choices to the expected frequencies under the null hypothesis that there exists no association between the independent and dependent variables.

The results of the Chi-squared tests are summarized as follows. The numbers in parentheses represent the \textit{degrees of freedom}, determined by the number of variable combinations. For example, with four privacy budgets and four mechanisms, the degrees of freedom are calculated as $df = (5 - 1) \times (4 - 1) = 4 \times 3 = 12$.

\begin{itemize}
    \itemsep 0em
    \item \textbf{Scenario}: \(\chi^2(4) = 48.51\), \(p < 0.001\). A significant relationship exists between the chosen privacy budget and the \textit{scenario} (\textit{Booking Chatbot} or \textit{ChatGPT}), suggesting that the context matters in making privacy decisions.
    \item \textbf{Sensitivity}: \(\chi^2(6) = 40.73\), \(p < 0.001\). There is a significant association between the chosen privacy budget and \textit{sensitivity}, showing that the perceived sensitivity of the text to be shared influences privacy selections.
    \item \textbf{Mechanism}: \(\chi^2(12) = 263.45\), \(p < 0.001\). There is a statistically significant relationship between the chosen privacy budget and \textit{mechanism}, indicating that the mechanism used affects the values selected by participants.
    \item \textbf{Reason}: \(\chi^2(6) = 19.08\), \(p < 0.05\). A significant relationship was found between the chosen privacy budget and \textit{reason}, indicating that the reason provided (specific, vague, or none) does affect the selected values.
\end{itemize}


\begin{table*}[htbp]
    \centering
    \begin{subtable}{0.27\linewidth}
        \centering
        \small
\resizebox{0.99\linewidth}{!}{        
\begin{tabular}{l|ccc}
 & \textbf{High} & \textbf{Low} & \textbf{None} \\ \hline
\textbf{High} & 1.0000 & \textbf{0.0054} &  \textbf{0.0000}\\
\textbf{Low} & \textbf{0.0054} & 1.0000 & 0.1951 \\
\textbf{None} & \textbf{0.0000} & 0.1951 & 1.0000
\end{tabular}
}
\caption{Sensitivity}
\label{tab:dunn_sensitivity}
    \end{subtable}
    \hfill
    \begin{subtable}{0.27\linewidth}
        \centering
        \resizebox{0.99\linewidth}{!}{    
\begin{tabular}{l|ccc}
 & \textbf{None} & \textbf{Specific} & \textbf{Vague} \\ \hline
\textbf{None} & 1.0000 & \textbf{0.0250} & 0.1971 \\
\textbf{Specific} & \textbf{0.0250} & 1.0000 & 1.0000 \\
\textbf{Vague} & 0.1971 & 1.0000 & 1.0000
\end{tabular}
}
\caption{Reason}
\label{tab:dunn_reason}
    \end{subtable}
    \hfill
    \begin{subtable}{0.38\linewidth}
    \centering
    \resizebox{0.99\linewidth}{!}{ 
\begin{tabular}{l|cccc}
 & \textbf{DP-BART} & \textbf{DP-MLM} & \textbf{DP-Prompt} & \textbf{TEM} \\ \hline
\textbf{DP-BART} & 1.0000 & \textbf{0.0000} & \textbf{0.0034} & \textbf{0.0000} \\
\textbf{DP-MLM} & \textbf{0.0000} & 1.0000 & \textbf{0.0000} & 1.0000 \\
\textbf{DP-Prompt} & \textbf{0.0034} & \textbf{0.0000} &  1.0000 & \textbf{0.0000} \\
\textbf{TEM} & \textbf{0.0000} & 1.0000 & \textbf{0.0000} & 1.0000
\end{tabular}
}
\caption{Mechanism}
\label{tab:dunn_mechanism}
    \end{subtable}
\caption{Dunn's post-hoc test results. \textbf{Bolded} $p$-values indicate statistically significant results ($p < 0.05$).}
\label{tab:dunn}
\end{table*}

\subsection{Hypothesis Testing}
As introduced in Section \ref{sec:hypo}, we posit that \textit{sensitivity} (H1), \textit{mechanism} (H2), and \textit{data collection reason} (H3) influence users' privacy choices in sharing their textual data. To test these hypotheses, we use a combination of Spearman's correlation \cite{spearman1904proof} and the Kruskal-Wallis H-test \cite{kruskal1952use}. The former allows us to analyze the correlation between the chosen privacy level and the factor in question (in the case of the ordinal \textit{sensitivity} and \textit{reason}), while the Kruskal-Wallis test informs us whether there exist any significant differences between the \textit{treatments} within these factors (e.g., \textit{High}, \textit{Low}, \textit{None} for \textit{sensitivity}). Additionally, we calculate the $\eta^2$ effect size\footnote{Given by $H$/($k - 1$), with $k$ as the group size.}, which gives an indication of the strength of the association. Finally, we perform a post-hoc Dunn's test \cite{dunn1964multiple} with Bonferroni correction\footnote{Multiplying each $p$-value by the total number of tests.}, which extends the analysis to explain between \textit{which} treatments there exist significant differences. The full Dunn's results are found in Table \ref{tab:dunn}.

\paragraph{H1.}
We calculate the following values to test for significance regarding H1:
\begin{itemize}
    \itemsep 0em
    \item \textbf{Spearman}: $\rho = -0.058, p < 0.001$
    \item \textbf{Kruskal-Wallis}: $H(2) = 24.83, p < 0.001$
    \item \textbf{Effect size ($\eta^2$)}: 0.0034
\end{itemize}
Thus, we observe a statistically significant correlation between the chosen privacy budget and sensitivity level. However, the effect size indicates that this correlation is quite weak. Dunn's post-hoc test reveals a statistically significant difference between \textit{High} and \textit{None} sensitivity ($p < 0.001$), but no significant difference involving \textit{Low}.

\paragraph{H2.}
We calculate the following values to test for significance regarding H2 (correlation not sensible here due to the categorical \textit{mechanism} variable):
\begin{itemize} 
    \itemsep 0em
    \item \textbf{Kruskal-Wallis}: $H(3) = 146.31, p < 0.001$
    \item \textbf{Effect size ($\eta^2$)}: 0.0203
\end{itemize}
We observe a significant difference in the selected privacy budget across our four selected mechanisms, supported by a \textit{small to medium effect} ($0.01 \le \eta^2 \le 0.06$). Dunn's post hoc reveals significant differences between both \textsc{TEM} and \textsc{DP-MLM} with both \textsc{DP-BART} and \textsc{DP-Prompt} (all with $p < 0.001$), showing a clear difference between generative and non-generative approaches. Additionally, a significant difference between \textsc{DP-BART} and \textsc{DP-Prompt} was observed ($p < 0.01$).

\paragraph{H3.}
We calculate the following values to test for significance regarding H3:
\begin{itemize}
    \itemsep 0em
    \item \textbf{Spearman}: $\rho = -0.030, p < 0.01$
    \item \textbf{Kruskal-Wallis}: $H(2) = 7.33, p < 0.05$
    \item \textbf{Effect size ($\eta^2$)}: 0.0010
\end{itemize}
Although this indicates significance, the effect size implies that providing a reason has little influence on the choice of privacy level. However, Dunn's post-hoc test shows a significant difference between \textit{specific} and \textit{none} ($p < 0.05$), but not between \textit{specific} and \textit{vague} or \textit{vague} and \textit{none}.

\section{Discussion}
In light of the presented findings, we reflect on the lessons learned and discuss their implications.

\paragraph{What Matters with Text Privatization.}
Our statistical analyses demonstrate that the factors of \textit{scenario}, \textit{sensitivity}, \textit{mechanism}, and \textit{reason} all play statistically significant roles in influencing a user's choices for text privatization, as indicated by the chi-square tests. However, these factors are not equally impactful, as we learn that the choice of DP mechanism is most important in swaying user perceptions of privacy options. In this, we provide empirical evidence that when dealing with natural language, it is also crucial \textit{how} text is privatized.

This above point is especially true in the case of text privatization with DP, where traditionally the $\varepsilon$ is seen as an arbitrator between privacy and utility. The insights we gain from our user study imply that deciding privacy budgets in deployed systems may not be as simple as \say{more privacy needed, then lower $\varepsilon$} and vice versa; instead, one must take into account the methods and context in which privatization is to occur. While this potentially makes the task of DP text privatization more challenging, it also provides more criteria by which researchers and practitioners can justify their privacy budgets.

\paragraph{Utility over Privacy?}
A very important finding regarding privatization preferences is manifested in the \textit{appearance} of private output texts. As can be seen in Figure \ref{fig:variables}, users were much more confident in choosing lower privacy budgets with the generative approaches (\textsc{DP-BART} and \textsc{DP-Prompt}), whereas \textsc{TEM} and \textsc{DP-MLM} received a significantly higher number of $\varepsilon = \infty$ choices. This suggests that when privatized texts are not as coherent or \say{natural} (as is in the case of word- or token-level, non-generative approaches), users tend to prefer coherence over privacy, a fact that seemingly contradicts the self-reported IUIPC privacy sentiments. Relating back to our choice of $\varepsilon$, these results point to a \say{tolerance} of at most 80\% cosine similarity (slider 4) or more, whereas lower values received far less selections. Such results imply an \say{acceptability range} for text privatization, which we observe to be somewhere between 80-100\% cosine similarity (this is of course specific to the chosen embedding model). In this, we learn that DP text privatization \textit{must} generate reasonable output texts before it will be more widely accepted. 

\paragraph{User Reasoning Patterns for Text Privatization.}
We analyzed and aggregated free-form survey feedback into four themes relating to privacy \say{reasoning patterns}. In particular, participants provided insights into \textit{why} they answered the way they did. For each pattern, we provide a representative quote.

\begin{itemize}
    \itemsep 0em
    \item \textbf{The need to find a balance}: \say{\textit{I tried to find the right balance between too much information and no information at all.}}
    \item \textbf{Depends on the use case}: \say{\textit{I felt more comfortable sharing my data with the medical booking platform than with ChatGPT, since I did not like the aspect of my data potentially being used for training their model.}}
    \item \textbf{Coherence is key}: \say{\textit{I chose the sentences which made the most sense written down. The other sentences on other points on the slider were not fully literate.}}
    \item \textbf{Personal information minimization}: \say{\textit{The less information given, the better.}}
\end{itemize}

Such patterns provide researchers with important insights into the thought processes of laypersons when reasoning about text privatization. Although some of these points may be quite challenging to realize technically, they set a framework for human-acceptable DP text privatization.

\paragraph{A Roadmap for DP NLP Research.}
The findings we present give way to a series of important factors that must be considered going forward:
\begin{enumerate}
    \itemsep 0em
    \item \textbf{DP NLP must be usable.} Focusing on text-to-text privatization with DP, we learn that well before other factors, the output of text privatization mechanisms must be coherent, correct, and readable; otherwise, perception of text privatization will be negative. Future work, therefore, would benefit from exploring what \textit{usability} in DP text privatization means.
    \item \textbf{DP NLP must consider context.} \textit{Context} here refers to factors beyond the technical privatization procedures: for what scenario is textual data collected or shared, what type(s) of personal information may be contained in the data, and perhaps to a lesser degree, for what purpose the data is meant. These factors affect what type of mechanism is needed, and moreover, how much \say{privacy} is required.
    \item \textbf{DP NLP must involve human studies.} Above all, our study teaches us that text privacy extends beyond technical challenges to the realm of \textit{socio}-technical challenges, such as increasing general user awareness and \textit{understanding} of how (DP) text privatization works and making clear what the implications of using such mechanisms are. Thus, we hold it crucial that further studies on \textit{usable} DP NLP not only extend our work, but also focus on designing methods for fostering acceptance of this promising, yet challenging technology.
\end{enumerate}

\section{Conclusion}
We conducted a survey study with 721 participants from six continents, investigating the influence of various factors on user perception of DP text privatization. Using a representative set of four DP mechanisms, we designed a series of vignettes to test for differences in the selection of text privatization level under a number of scenarios. We found that all tested factors play an important role in the context of text privatization, yet the factor of mechanism design is the most salient. In particular, mechanisms producing clearer and more natural outputs encourage users to choose higher privacy levels (lower $\varepsilon$ budgets). Our findings reveal the importance of involving the general population in guiding the direction of DP NLP research, and we hope that our work motivates future studies on aligning DP NLP research and practice.

\newpage
\section*{Acknowledgements}
We would like to thank the anonymous reviewers for their feedback, as well as the committee and all survey participants for their valuable contributions. We also greatly appreciate Jian Kong and Timo Kühne for their assistance in survey deployment.

\section*{Limitations}
The primary limitation of our work is inherent to conducting a general user study addressing a complex technical topic, such as DP text privatization. Although we focused on clear and understandable instructions for survey participants, we cannot be certain that all participants fully understood the task at hand. Indeed, in the feedback section, we received a number of comments with users expressing concern that they did not fully understand the task; while such comments were in the vast minority, this could still affect the calculation of our results. Nevertheless, we mitigated this threat to validity by selecting a large sample size, where each of the 72 vignettes was answered by at least 100 survey participants. We hope that future works will alleviate this challenge by working on standardized methods for communicating DP NLP topics.

Another clear limitation relates to the choice of four mechanisms that served as the basis for the vignettes we designed for the surveys. We did not perform any cleanup or post-processing of the privatized texts, often resulting in obvious grammatical errors (in the case of word-level privatization) or non-ASCII characters (in the case of the generative approaches), which could plausibly have biased the selection of slider values in the survey. While this was difficult to avoid, we argue that this enabled insights regarding different perceptions of different mechanism outputs, leading us to the conclusion that this factor is of utmost importance.

Finally, we caution that our survey sample may not be entirely representative in terms of the global population and language domains. The use of the Prolific platform limited our control over survey population, resulting in a particularly high number of respondents from South Africa, while less representation was had from North and South America and Asia. Furthermore, we perform no analyses regarding differences across regions, genders, professions, or educational backgrounds. Additionally, the primary focus was on texts related to the \textit{medical} domain, as a result of the selected texts from our committee vote. Ideally, future studies could replicate our findings given different, more representative samples and broader text domains.

\section*{Ethics Statement}
Our study was reviewed and approved by the ethics commission of the Technical University of Munich, with approval number 2024-86-NM-BA.

Particularly regarding the involvement of human subjects in our study, we affirm that participation was completely voluntary and compensated with a fair wage via the Prolific platform. Outside of the initial pilot study, no preference was given to any potential survey participant; this was conducted on a first-come-first-serve basis facilitated by Prolific. We ensured the well-being of our participants by creating an inviting and easy-to-navigate survey application, engaging (anonymously) with participants who had questions or concerns during the survey conduction, and not collecting or storing any personally identifiable information at any point. 

As this work is centered on the timely and important topic of data privacy, we hope that its impact extends to both researchers working in the field of privacy (in NLP), as well as to end users who may increase their knowledge and awareness of current trends and issues in privacy research. In particular, we envision similar types of studies becoming more commonplace in privacy-preserving NLP research, and we hope that this work contributes positively to motivating such future works.


\bibliography{custom}

\begin{thebibliography}{59}
\providecommand{\natexlab}[1]{#1}

\bibitem[{Arnold et~al.(2023{\natexlab{a}})Arnold, Yesilbas, and Weinzierl}]{arnold-etal-2023-driving}
Stefan Arnold, Dilara Yesilbas, and Sven Weinzierl. 2023{\natexlab{a}}.
\newblock \href {https://doi.org/10.18653/v1/2023.trustnlp-1.2} {Driving context into text-to-text privatization}.
\newblock In \emph{Proceedings of the 3rd Workshop on Trustworthy Natural Language Processing (TrustNLP 2023)}, pages 15--25, Toronto, Canada. Association for Computational Linguistics.

\bibitem[{Arnold et~al.(2023{\natexlab{b}})Arnold, Yesilbas, and Weinzierl}]{arnold-etal-2023-guiding}
Stefan Arnold, Dilara Yesilbas, and Sven Weinzierl. 2023{\natexlab{b}}.
\newblock \href {https://doi.org/10.18653/v1/2023.trustnlp-1.14} {Guiding text-to-text privatization by syntax}.
\newblock In \emph{Proceedings of the 3rd Workshop on Trustworthy Natural Language Processing (TrustNLP 2023)}, pages 151--162, Toronto, Canada. Association for Computational Linguistics.

\bibitem[{Atzm{\"u}ller and Steiner(2010)}]{atzmuller2010experimental}
Christiane Atzm{\"u}ller and Peter~M Steiner. 2010.
\newblock \href {https://doi.org/10.1027/1614-2241/a000014} {Experimental vignette studies in survey research}.
\newblock \emph{Methodology}.

\bibitem[{Auspurg and Hinz(2014)}]{auspurg2014factorial}
Katrin Auspurg and Thomas Hinz. 2014.
\newblock \href {https://doi.org/10.4135/9781483398075} {\emph{Factorial survey experiments}}, volume 175.
\newblock SAGE Publications, Inc.

\bibitem[{Beigi et~al.(2019)Beigi, Shu, Guo, Wang, and Liu}]{10.1145/3342220.3344925}
Ghazaleh Beigi, Kai Shu, Ruocheng Guo, Suhang Wang, and Huan Liu. 2019.
\newblock \href {https://doi.org/10.1145/3342220.3344925} {Privacy preserving text representation learning}.
\newblock In \emph{Proceedings of the 30th ACM Conference on Hypertext and Social Media}, HT '19, page 275–276, New York, NY, USA. Association for Computing Machinery.

\bibitem[{Bo et~al.(2021)Bo, Ding, Fung, and Iqbal}]{bo-etal-2021-er}
Haohan Bo, Steven H.~H. Ding, Benjamin C.~M. Fung, and Farkhund Iqbal. 2021.
\newblock \href {https://doi.org/10.18653/v1/2021.naacl-main.314} {{ER}-{AE}: Differentially private text generation for authorship anonymization}.
\newblock In \emph{Proceedings of the 2021 Conference of the North American Chapter of the Association for Computational Linguistics: Human Language Technologies}, pages 3997--4007, Online. Association for Computational Linguistics.

\bibitem[{Bollegala et~al.(2023)Bollegala, Otake, Machide, and Kawarabayashi}]{bollegala-etal-2023-neighbourhood}
Danushka Bollegala, Shuichi Otake, Tomoya Machide, and Ken-ichi Kawarabayashi. 2023.
\newblock \href {https://doi.org/10.18653/v1/2023.findings-ijcnlp.7} {A neighbourhood-aware differential privacy mechanism for static word embeddings}.
\newblock In \emph{Findings of the Association for Computational Linguistics: IJCNLP-AACL 2023 (Findings)}, pages 65--79, Nusa Dua, Bali. Association for Computational Linguistics.

\bibitem[{Carvalho et~al.(2021)Carvalho, Vasiloudis, and Feyisetan}]{carvalho2021brrpreservingprivacytext}
Ricardo~Silva Carvalho, Theodore Vasiloudis, and Oluwaseyi Feyisetan. 2021.
\newblock \href {https://arxiv.org/abs/2107.07923} {{BRR}: Preserving privacy of text data efficiently on device}.
\newblock \emph{Preprint}, arXiv:2107.07923.

\bibitem[{Carvalho et~al.(2023)Carvalho, Vasiloudis, Feyisetan, and Wang}]{carvalho2023tem}
Ricardo~Silva Carvalho, Theodore Vasiloudis, Oluwaseyi Feyisetan, and Ke~Wang. 2023.
\newblock \href {https://doi.org/10.1137/1.9781611977653.ch99} {{TEM}: High utility metric differential privacy on text}.
\newblock In \emph{Proceedings of the 2023 SIAM International Conference on Data Mining (SDM)}, pages 883--890. SIAM.

\bibitem[{Chen et~al.(2023)Chen, Mo, Wang, Chen, Nie, Wang, and Cui}]{Chen2022ACT}
Sai Chen, Fengran Mo, Yanhao Wang, Cen Chen, Jian-Yun Nie, Chengyu Wang, and Jamie Cui. 2023.
\newblock \href {https://doi.org/10.18653/v1/2023.findings-acl.355} {A customized text sanitization mechanism with differential privacy}.
\newblock In \emph{Findings of the Association for Computational Linguistics: ACL 2023}, pages 5747--5758, Toronto, Canada. Association for Computational Linguistics.

\bibitem[{Cummings et~al.(2021)Cummings, Kaptchuk, and Redmiles}]{10.1145/3460120.3485252}
Rachel Cummings, Gabriel Kaptchuk, and Elissa~M. Redmiles. 2021.
\newblock \href {https://doi.org/10.1145/3460120.3485252} {"{I} need a better description": An investigation into user expectations for differential privacy}.
\newblock In \emph{Proceedings of the 2021 ACM SIGSAC Conference on Computer and Communications Security}, CCS '21, page 3037–3052, New York, NY, USA. Association for Computing Machinery.

\bibitem[{Du et~al.(2023)Du, Yue, Chow, and Sun}]{10.1145/3543507.3583512}
Minxin Du, Xiang Yue, Sherman S.~M. Chow, and Huan Sun. 2023.
\newblock \href {https://doi.org/10.1145/3543507.3583512} {Sanitizing sentence embeddings (and labels) for local differential privacy}.
\newblock In \emph{Proceedings of the ACM Web Conference 2023}, WWW '23, page 2349–2359, New York, NY, USA. Association for Computing Machinery.

\bibitem[{Dunn(1964)}]{dunn1964multiple}
Olive~Jean Dunn. 1964.
\newblock \href {https://doi.org/10.1080/00401706.1964.10490181} {Multiple comparisons using rank sums}.
\newblock \emph{Technometrics}, 6(3):241--252.

\bibitem[{Evans et~al.(2015)Evans, Roberts, Keeley, Blossom, Amaro, Garcia, Stough, Canter, Robles, and Reed}]{EVANS2015160}
Spencer~C. Evans, Michael~C. Roberts, Jared~W. Keeley, Jennifer~B. Blossom, Christina~M. Amaro, Andrea~M. Garcia, Cathleen~Odar Stough, Kimberly~S. Canter, Rebeca Robles, and Geoffrey~M. Reed. 2015.
\newblock \href {https://doi.org/10.1016/j.ijchp.2014.12.001} {Vignette methodologies for studying clinicians’ decision-making: Validity, utility, and application in icd-11 field studies}.
\newblock \emph{International Journal of Clinical and Health Psychology}, 15(2):160--170.

\bibitem[{Fernandes et~al.(2019)Fernandes, Dras, and McIver}]{fernandes2019generalised}
Natasha Fernandes, Mark Dras, and Annabelle McIver. 2019.
\newblock \href {https://doi.org/10.1007/978-3-030-17138-4_6} {Generalised differential privacy for text document processing}.
\newblock In \emph{Principles of Security and Trust: 8th International Conference, POST 2019, Held as Part of the European Joint Conferences on Theory and Practice of Software, ETAPS 2019, Prague, Czech Republic, April 6--11, 2019, Proceedings 8}, pages 123--148. Springer International Publishing.

\bibitem[{Feyisetan et~al.(2020)Feyisetan, Balle, Drake, and Diethe}]{10.1145/3336191.3371856}
Oluwaseyi Feyisetan, Borja Balle, Thomas Drake, and Tom Diethe. 2020.
\newblock \href {https://doi.org/10.1145/3336191.3371856} {Privacy- and utility-preserving textual analysis via calibrated multivariate perturbations}.
\newblock In \emph{Proceedings of the 13th International Conference on Web Search and Data Mining}, WSDM '20, page 178–186, New York, NY, USA. Association for Computing Machinery.

\bibitem[{Feyisetan et~al.(2019)Feyisetan, Diethe, and Drake}]{8970912}
Oluwaseyi Feyisetan, Tom Diethe, and Thomas Drake. 2019.
\newblock \href {https://doi.org/10.1109/ICDM.2019.00031} {Leveraging hierarchical representations for preserving privacy and utility in text}.
\newblock In \emph{2019 IEEE International Conference on Data Mining (ICDM)}, pages 210--219.

\bibitem[{Feyisetan and Kasiviswanathan(2021)}]{feyisetan-kasiviswanathan-2021-private}
Oluwaseyi Feyisetan and Shiva Kasiviswanathan. 2021.
\newblock \href {https://doi.org/10.18653/v1/2021.trustnlp-1.3} {Private release of text embedding vectors}.
\newblock In \emph{Proceedings of the First Workshop on Trustworthy Natural Language Processing}, pages 15--27, Online. Association for Computational Linguistics.

\bibitem[{Franzen et~al.(2022)Franzen, Nu\~{n}ez~von Voigt, S\"{o}rries, Tschorsch, and M\"{u}ller-Birn}]{10.1145/3548606.3560693}
Daniel Franzen, Saskia Nu\~{n}ez~von Voigt, Peter S\"{o}rries, Florian Tschorsch, and Claudia M\"{u}ller-Birn. 2022.
\newblock \href {https://doi.org/10.1145/3548606.3560693} {Am i private and if so, how many? communicating privacy guarantees of differential privacy with risk communication formats}.
\newblock In \emph{Proceedings of the 2022 ACM SIGSAC Conference on Computer and Communications Security}, CCS '22, page 1125–1139, New York, NY, USA. Association for Computing Machinery.

\bibitem[{Gro{\ss}(2021)}]{gross2021validity}
Thomas Gro{\ss}. 2021.
\newblock \href {https://doi.org/10.2478/popets-2021-0026} {Validity and reliability of the scale internet users’ information privacy concerns (iuipc)}.
\newblock \emph{Proceedings on Privacy Enhancing Technologies}.

\bibitem[{Habernal(2021)}]{habernal-2021-differential}
Ivan Habernal. 2021.
\newblock \href {https://doi.org/10.18653/v1/2021.emnlp-main.114} {When differential privacy meets {NLP}: The devil is in the detail}.
\newblock In \emph{Proceedings of the 2021 Conference on Empirical Methods in Natural Language Processing}, pages 1522--1528, Online and Punta Cana, Dominican Republic. Association for Computational Linguistics.

\bibitem[{Hu et~al.(2024)Hu, Habernal, Shen, and Wang}]{hu-etal-2024-differentially}
Lijie Hu, Ivan Habernal, Lei Shen, and Di~Wang. 2024.
\newblock \href {https://aclanthology.org/2024.findings-eacl.33} {Differentially private natural language models: Recent advances and future directions}.
\newblock In \emph{Findings of the Association for Computational Linguistics: EACL 2024}, pages 478--499, St. Julian{'}s, Malta. Association for Computational Linguistics.

\bibitem[{Igamberdiev et~al.(2022)Igamberdiev, Arnold, and Habernal}]{igamberdiev-etal-2022-dp}
Timour Igamberdiev, Thomas Arnold, and Ivan Habernal. 2022.
\newblock \href {https://aclanthology.org/2022.coling-1.258} {{DP}-rewrite: Towards reproducibility and transparency in differentially private text rewriting}.
\newblock In \emph{Proceedings of the 29th International Conference on Computational Linguistics}, pages 2927--2933, Gyeongju, Republic of Korea. International Committee on Computational Linguistics.

\bibitem[{Igamberdiev and Habernal(2023)}]{igamberdiev-habernal-2023-dp}
Timour Igamberdiev and Ivan Habernal. 2023.
\newblock \href {https://doi.org/10.18653/v1/2023.findings-acl.874} {{DP}-{BART} for privatized text rewriting under local differential privacy}.
\newblock In \emph{Findings of the Association for Computational Linguistics: ACL 2023}, pages 13914--13934, Toronto, Canada. Association for Computational Linguistics.

\bibitem[{Igamberdiev et~al.(2024)Igamberdiev, Vu, Kuennecke, Yu, Holmer, and Habernal}]{igamberdiev-etal-2024-dp}
Timour Igamberdiev, Doan Nam~Long Vu, Felix Kuennecke, Zhuo Yu, Jannik Holmer, and Ivan Habernal. 2024.
\newblock \href {https://aclanthology.org/2024.eacl-demo.11/} {{DP}-{NMT}: Scalable differentially private machine translation}.
\newblock In \emph{Proceedings of the 18th Conference of the European Chapter of the Association for Computational Linguistics: System Demonstrations}, pages 94--105, St. Julians, Malta. Association for Computational Linguistics.

\bibitem[{Imola et~al.(2022)Imola, Kasiviswanathan, White, Aggarwal, and Teissier}]{Imola2022}
Jacob Imola, Shiva Kasiviswanathan, Stephen White, Abhinav Aggarwal, and Nathanael Teissier. 2022.
\newblock \href {https://www.amazon.science/publications/balancing-utility-and-scalability-in-metric-differential-privacy} {Balancing utility and scalability in metric differential privacy}.
\newblock In \emph{The 38th Conference on Uncertainty in Artificial Intelligence}.

\bibitem[{Karegar et~al.(2022)Karegar, Alaqra, and Fischer-H{\"u}bner}]{281270}
Farzaneh Karegar, Ala~Sarah Alaqra, and Simone Fischer-H{\"u}bner. 2022.
\newblock \href {https://www.usenix.org/conference/soups2022/presentation/karegar} {Exploring {User-Suitable} metaphors for differentially private data analyses}.
\newblock In \emph{Eighteenth Symposium on Usable Privacy and Security (SOUPS 2022)}, pages 175--193, Boston, MA. USENIX Association.

\bibitem[{Klymenko et~al.(2022)Klymenko, Meisenbacher, and Matthes}]{klymenko-etal-2022-differential}
Oleksandra Klymenko, Stephen Meisenbacher, and Florian Matthes. 2022.
\newblock \href {https://doi.org/10.18653/v1/2022.privatenlp-1.1} {Differential privacy in natural language processing: The story so far}.
\newblock In \emph{Proceedings of the Fourth Workshop on Privacy in Natural Language Processing}, pages 1--11, Seattle, United States. Association for Computational Linguistics.

\bibitem[{Krishna et~al.(2021)Krishna, Gupta, and Dupuy}]{krishna-etal-2021-adept}
Satyapriya Krishna, Rahul Gupta, and Christophe Dupuy. 2021.
\newblock \href {https://doi.org/10.18653/v1/2021.eacl-main.207} {{AD}e{PT}: Auto-encoder based differentially private text transformation}.
\newblock In \emph{Proceedings of the 16th Conference of the European Chapter of the Association for Computational Linguistics: Main Volume}, pages 2435--2439, Online. Association for Computational Linguistics.

\bibitem[{Kruskal and Wallis(1952)}]{kruskal1952use}
William~H Kruskal and W~Allen Wallis. 1952.
\newblock \href {https://doi.org/10.1080/01621459.1952.10483441} {Use of ranks in one-criterion variance analysis}.
\newblock \emph{Journal of the American statistical Association}, 47(260):583--621.

\bibitem[{Lewis et~al.(2020)Lewis, Liu, Goyal, Ghazvininejad, Mohamed, Levy, Stoyanov, and Zettlemoyer}]{lewis-etal-2020-bart}
Mike Lewis, Yinhan Liu, Naman Goyal, Marjan Ghazvininejad, Abdelrahman Mohamed, Omer Levy, Veselin Stoyanov, and Luke Zettlemoyer. 2020.
\newblock \href {https://doi.org/10.18653/v1/2020.acl-main.703} {{BART}: Denoising sequence-to-sequence pre-training for natural language generation, translation, and comprehension}.
\newblock In \emph{Proceedings of the 58th Annual Meeting of the Association for Computational Linguistics}, pages 7871--7880, Online. Association for Computational Linguistics.

\bibitem[{Lyu et~al.(2020{\natexlab{a}})Lyu, He, and Li}]{lyu-etal-2020-differentially}
Lingjuan Lyu, Xuanli He, and Yitong Li. 2020{\natexlab{a}}.
\newblock \href {https://doi.org/10.18653/v1/2020.findings-emnlp.213} {Differentially private representation for {NLP}: Formal guarantee and an empirical study on privacy and fairness}.
\newblock In \emph{Findings of the Association for Computational Linguistics: EMNLP 2020}, pages 2355--2365, Online. Association for Computational Linguistics.

\bibitem[{Lyu et~al.(2020{\natexlab{b}})Lyu, Li, He, and Xiao}]{10.1145/3397271.3401260}
Lingjuan Lyu, Yitong Li, Xuanli He, and Tong Xiao. 2020{\natexlab{b}}.
\newblock \href {https://doi.org/10.1145/3397271.3401260} {Towards differentially private text representations}.
\newblock In \emph{Proceedings of the 43rd International ACM SIGIR Conference on Research and Development in Information Retrieval}, SIGIR '20, page 1813–1816, New York, NY, USA. Association for Computing Machinery.

\bibitem[{Maheshwari et~al.(2022)Maheshwari, Denis, Keller, and Bellet}]{maheshwari-etal-2022-fair}
Gaurav Maheshwari, Pascal Denis, Mikaela Keller, and Aur{\'e}lien Bellet. 2022.
\newblock \href {https://doi.org/10.18653/v1/2022.findings-emnlp.514} {Fair {NLP} models with differentially private text encoders}.
\newblock In \emph{Findings of the Association for Computational Linguistics: EMNLP 2022}, pages 6913--6930, Abu Dhabi, United Arab Emirates. Association for Computational Linguistics.

\bibitem[{Malhotra et~al.(2004)Malhotra, Kim, and Agarwal}]{malhotra2004internet}
Naresh~K Malhotra, Sung~S Kim, and James Agarwal. 2004.
\newblock \href {https://doi.org/10.1287/isre.1040.0032} {Internet users' information privacy concerns (iuipc): The construct, the scale, and a causal model}.
\newblock \emph{Information systems research}, 15(4):336--355.

\bibitem[{Mattern et~al.(2022{\natexlab{a}})Mattern, Jin, Weggenmann, Schoelkopf, and Sachan}]{mattern-etal-2022-differentially}
Justus Mattern, Zhijing Jin, Benjamin Weggenmann, Bernhard Schoelkopf, and Mrinmaya Sachan. 2022{\natexlab{a}}.
\newblock \href {https://doi.org/10.18653/v1/2022.emnlp-main.323} {Differentially private language models for secure data sharing}.
\newblock In \emph{Proceedings of the 2022 Conference on Empirical Methods in Natural Language Processing}, pages 4860--4873, Abu Dhabi, United Arab Emirates. Association for Computational Linguistics.

\bibitem[{Mattern et~al.(2022{\natexlab{b}})Mattern, Weggenmann, and Kerschbaum}]{mattern-etal-2022-limits}
Justus Mattern, Benjamin Weggenmann, and Florian Kerschbaum. 2022{\natexlab{b}}.
\newblock \href {https://doi.org/10.18653/v1/2022.findings-naacl.65} {The limits of word level differential privacy}.
\newblock In \emph{Findings of the Association for Computational Linguistics: NAACL 2022}, pages 867--881, Seattle, United States. Association for Computational Linguistics.

\bibitem[{Meehan et~al.(2022)Meehan, Mrini, and Chaudhuri}]{meehan-etal-2022-sentence}
Casey Meehan, Khalil Mrini, and Kamalika Chaudhuri. 2022.
\newblock \href {https://doi.org/10.18653/v1/2022.acl-long.238} {Sentence-level privacy for document embeddings}.
\newblock In \emph{Proceedings of the 60th Annual Meeting of the Association for Computational Linguistics (Volume 1: Long Papers)}, pages 3367--3380, Dublin, Ireland. Association for Computational Linguistics.

\bibitem[{Meisenbacher et~al.(2024{\natexlab{a}})Meisenbacher, Chevli, Vladika, and Matthes}]{meisenbacher-etal-2024-dp}
Stephen Meisenbacher, Maulik Chevli, Juraj Vladika, and Florian Matthes. 2024{\natexlab{a}}.
\newblock \href {https://doi.org/10.18653/v1/2024.findings-acl.554} {{DP}-{MLM}: Differentially private text rewriting using masked language models}.
\newblock In \emph{Findings of the Association for Computational Linguistics: ACL 2024}, pages 9314--9328, Bangkok, Thailand. Association for Computational Linguistics.

\bibitem[{Meisenbacher and Matthes(2024{\natexlab{a}})}]{10.1145/3664476.3669926}
Stephen Meisenbacher and Florian Matthes. 2024{\natexlab{a}}.
\newblock \href {https://doi.org/10.1145/3664476.3669926} {Just rewrite it again: A post-processing method for enhanced semantic similarity and privacy preservation of differentially private rewritten text}.
\newblock In \emph{Proceedings of the 19th International Conference on Availability, Reliability and Security}, ARES '24, New York, NY, USA. Association for Computing Machinery.

\bibitem[{Meisenbacher and Matthes(2024{\natexlab{b}})}]{meisenbacher-matthes-2024-thinking}
Stephen Meisenbacher and Florian Matthes. 2024{\natexlab{b}}.
\newblock \href {https://doi.org/10.18653/v1/2024.emnlp-main.324} {Thinking outside of the differential privacy box: A case study in text privatization with language model prompting}.
\newblock In \emph{Proceedings of the 2024 Conference on Empirical Methods in Natural Language Processing}, pages 5656--5665, Miami, Florida, USA. Association for Computational Linguistics.

\bibitem[{Meisenbacher et~al.(2024{\natexlab{b}})Meisenbacher, Nandakumar, Klymenko, and Matthes}]{meisenbacher2024comparative}
Stephen Meisenbacher, Nihildev Nandakumar, Alexandra Klymenko, and Florian Matthes. 2024{\natexlab{b}}.
\newblock \href {https://aclanthology.org/2024.lrec-main.16} {A comparative analysis of word-level metric differential privacy: Benchmarking the privacy-utility trade-off}.
\newblock In \emph{Proceedings of the 2024 Joint International Conference on Computational Linguistics, Language Resources and Evaluation (LREC-COLING 2024)}, pages 174--185, Torino, Italia. ELRA and ICCL.

\bibitem[{Munn et~al.(2018)Munn, Peters, Stern, Tufanaru, McArthur, and Aromataris}]{munn2018systematic}
Zachary Munn, Micah~DJ Peters, Cindy Stern, Catalin Tufanaru, Alexa McArthur, and Edoardo Aromataris. 2018.
\newblock \href {https://doi.org/10.1186/s12874-018-0611-x} {Systematic review or scoping review? guidance for authors when choosing between a systematic or scoping review approach}.
\newblock \emph{BMC medical research methodology}, 18:1--7.

\bibitem[{Nanayakkara et~al.(2023)Nanayakkara, Smart, Cummings, Kaptchuk, and Redmiles}]{10.5555/3620237.3620328}
Priyanka Nanayakkara, Mary~Anne Smart, Rachel Cummings, Gabriel Kaptchuk, and Elissa~M. Redmiles. 2023.
\newblock \href {https://www.usenix.org/conference/usenixsecurity23/presentation/nanayakkara} {What are the chances? explaining the epsilon parameter in differential privacy}.
\newblock In \emph{Proceedings of the 32nd USENIX Conference on Security Symposium}, SEC '23, USA. USENIX Association.

\bibitem[{Plant et~al.(2021)Plant, Gkatzia, and Giuffrida}]{plant-etal-2021-cape}
Richard Plant, Dimitra Gkatzia, and Valerio Giuffrida. 2021.
\newblock \href {https://doi.org/10.18653/v1/2021.emnlp-main.628} {{CAPE}: Context-aware private embeddings for private language learning}.
\newblock In \emph{Proceedings of the 2021 Conference on Empirical Methods in Natural Language Processing}, pages 7970--7978, Online and Punta Cana, Dominican Republic. Association for Computational Linguistics.

\bibitem[{Reimers and Gurevych(2019)}]{reimers-gurevych-2019-sentence}
Nils Reimers and Iryna Gurevych. 2019.
\newblock \href {https://doi.org/10.18653/v1/D19-1410} {Sentence-{BERT}: Sentence embeddings using {S}iamese {BERT}-networks}.
\newblock In \emph{Proceedings of the 2019 Conference on Empirical Methods in Natural Language Processing and the 9th International Joint Conference on Natural Language Processing (EMNLP-IJCNLP)}, pages 3982--3992, Hong Kong, China. Association for Computational Linguistics.

\bibitem[{Shapiro and Wilk(1965)}]{shapiro1965analysis}
Samuel~Sanford Shapiro and Martin~B Wilk. 1965.
\newblock \href {https://doi.org/10.1093/biomet/52.3-4.591} {An analysis of variance test for normality (complete samples)}.
\newblock \emph{Biometrika}, 52(3-4):591--611.

\bibitem[{Smart et~al.(2022)Smart, Sood, and Vaccaro}]{10.1145/3555762}
Mary~Anne Smart, Dhruv Sood, and Kristen Vaccaro. 2022.
\newblock \href {https://doi.org/10.1145/3555762} {Understanding risks of privacy theater with differential privacy}.
\newblock \emph{Proc. ACM Hum.-Comput. Interact.}, 6(CSCW2).

\bibitem[{Spearman(1904)}]{spearman1904proof}
C~Spearman. 1904.
\newblock \href {https://doi.org/10.2307/1412159} {The proof and measurement of association between two things}.
\newblock \emph{The American Journal of Psychology}, 15(1):72--101.

\bibitem[{Utpala et~al.(2023)Utpala, Hooker, and Chen}]{utpala-etal-2023-locally}
Saiteja Utpala, Sara Hooker, and Pin-Yu Chen. 2023.
\newblock \href {https://doi.org/10.18653/v1/2023.findings-emnlp.566} {Locally differentially private document generation using zero shot prompting}.
\newblock In \emph{Findings of the Association for Computational Linguistics: EMNLP 2023}, pages 8442--8457, Singapore. Association for Computational Linguistics.

\bibitem[{Vu et~al.(2024)Vu, Igamberdiev, and Habernal}]{vu-etal-2024-granularity}
Doan Nam~Long Vu, Timour Igamberdiev, and Ivan Habernal. 2024.
\newblock \href {https://doi.org/10.18653/v1/2024.findings-emnlp.29} {Granularity is crucial when applying differential privacy to text: An investigation for neural machine translation}.
\newblock In \emph{Findings of the Association for Computational Linguistics: EMNLP 2024}, pages 507--527, Miami, Florida, USA. Association for Computational Linguistics.

\bibitem[{Weggenmann and Kerschbaum(2018)}]{weggenmann2018syntf}
Benjamin Weggenmann and Florian Kerschbaum. 2018.
\newblock \href {https://doi.org/10.1145/3209978.3210008} {Syn{TF}: Synthetic and differentially private term frequency vectors for privacy-preserving text mining}.
\newblock In \emph{The 41st International ACM SIGIR Conference on Research \& Development in Information Retrieval}, pages 305--314.

\bibitem[{Weggenmann et~al.(2022)Weggenmann, Rublack, Andrejczuk, Mattern, and Kerschbaum}]{10.1145/3485447.3512232}
Benjamin Weggenmann, Valentin Rublack, Michael Andrejczuk, Justus Mattern, and Florian Kerschbaum. 2022.
\newblock \href {https://doi.org/10.1145/3485447.3512232} {{DP-VAE}: Human-readable text anonymization for online reviews with differentially private variational autoencoders}.
\newblock In \emph{Proceedings of the ACM Web Conference 2022}, WWW '22, page 721–731, New York, NY, USA. Association for Computing Machinery.

\bibitem[{Weiss et~al.(2024)Weiss, Kreuter, and Habernal}]{weiss-etal-2024-share}
Christopher Weiss, Frauke Kreuter, and Ivan Habernal. 2024.
\newblock \href {https://aclanthology.org/2024.lrec-main.1419/} {To share or not to share: What risks would laypeople accept to give sensitive data to differentially-private {NLP} systems?}
\newblock In \emph{Proceedings of the 2024 Joint International Conference on Computational Linguistics, Language Resources and Evaluation (LREC-COLING 2024)}, pages 16331--16342, Torino, Italia. ELRA and ICCL.

\bibitem[{Xiong et~al.(2020)Xiong, Wang, Li, and Jha}]{xiong2020towards}
Aiping Xiong, Tianhao Wang, Ninghui Li, and Somesh Jha. 2020.
\newblock \href {https://doi.org/10.1109/SP40000.2020.00088} {Towards effective differential privacy communication for users’ data sharing decision and comprehension}.
\newblock In \emph{2020 IEEE Symposium on Security and Privacy (SP)}, pages 392--410. IEEE.

\bibitem[{Xu et~al.(2021{\natexlab{a}})Xu, Feyisetan, Aggarwal, Xu, and Teissier}]{xu2021density}
Nan Xu, Oluwaseyi Feyisetan, Abhinav Aggarwal, Zekun Xu, and Nathanael Teissier. 2021{\natexlab{a}}.
\newblock \href {https://doi.org/10.32473/flairs.v34i1.128463} {Density-aware differentially private textual perturbations using truncated gumbel noise}.
\newblock In \emph{The International FLAIRS Conference Proceedings}, volume~34.

\bibitem[{Xu et~al.(2020)Xu, Aggarwal, Feyisetan, and Teissier}]{xu-etal-2020-differentially}
Zekun Xu, Abhinav Aggarwal, Oluwaseyi Feyisetan, and Nathanael Teissier. 2020.
\newblock \href {https://doi.org/10.18653/v1/2020.privatenlp-1.2} {A differentially private text perturbation method using regularized mahalanobis metric}.
\newblock In \emph{Proceedings of the Second Workshop on Privacy in NLP}, pages 7--17, Online. Association for Computational Linguistics.

\bibitem[{Xu et~al.(2021{\natexlab{b}})Xu, Aggarwal, Feyisetan, and Teissier}]{xu-etal-2021-utilitarian}
Zekun Xu, Abhinav Aggarwal, Oluwaseyi Feyisetan, and Nathanael Teissier. 2021{\natexlab{b}}.
\newblock \href {https://doi.org/10.18653/v1/2021.privatenlp-1.2} {On a utilitarian approach to privacy preserving text generation}.
\newblock In \emph{Proceedings of the Third Workshop on Privacy in Natural Language Processing}, pages 11--20, Online. Association for Computational Linguistics.

\bibitem[{Yue et~al.(2021)Yue, Du, Wang, Li, Sun, and Chow}]{yue}
Xiang Yue, Minxin Du, Tianhao Wang, Yaliang Li, Huan Sun, and Sherman S.~M. Chow. 2021.
\newblock \href {https://doi.org/10.18653/v1/2021.findings-acl.337} {Differential privacy for text analytics via natural text sanitization}.
\newblock In \emph{Findings of the Association for Computational Linguistics: ACL-IJCNLP 2021}, pages 3853--3866, Online. Association for Computational Linguistics.

\end{thebibliography}

\newpage
\appendix
\onecolumn


\newpage
\section{Sensitive Data Sharing Scenarios -- Committee Vote}
\label{sec:vote}
{\scriptsize
Thank you for taking the time to participate in this survey.

Background: we are well on our way in a study investigating user perceptions of (text) data sharing. Specifically, we aim to study the effect of Differential Privacy (DP) rewriting mechanisms, more particularly the effect of the privacy parameter (epsilon). 

For the research, we have opted to conduct a vignette study, in which users will be prompted to place themselves into a provided scenario, thereafter answering to what extent they are comfortable sharing their text data given different levels of text privatization.

In the administered survey, we plan on presenting two overarching vignettes (with varying parameters, but not important here). To start, we have drafted a number of such vignettes, with the goal of narrowing down to the two most \textbf{relevant, plausible, and understandable} scenarios. For this, we need your help!

Please answer the following questions to the best of your ability. By doing so, you are helping to advance our study. Welcome to the committee :)

\hrulefill

\textbf{Candidate Vignettes}
\newline

For each of the following candidates, you will be asked how well the scenario depicts a \say{sensitive} data sharing scenario. As introduced, we are searching for the best vignettes in terms of:
\begin{itemize}
    \item \textbf{Relevance}: this is a timely and relevant scenario, and it is a scenario which indeed involves some sensitive or private information.
    \item \textbf{Plausibility}: this is something which you can imagine actually taking place in the real world. It does not have to be exactly so.
    \item \textbf{Understandability}: the scenario makes sense to you -- there are no major ambiguities as to what is going on.
\end{itemize}
For each question, you will first be presented with a textual description of the scenario. There are three levels of \say{sensitive information}, corresponding to three versions of the vignette, so it is important to view the scenarios as a whole. You will then be asked to judge how well this scenario overall fits the above criteria.

\hrulefill

\textbf{Candidate 1a}
\newline

Bob is researching his health condition with ChatGPT, and he types the following message to the chatbot: 

\begin{itemize}
    \item[a)] \textbf{Highly Sensitive Information}: \say{I was diagnosed with a lung cancer last week and I’m feeling overwhelmed. I’ll be treated at IsarHealth in Munich starting June 1st. Can you tell me some information about treatments and potential side effects?} 
    \item[b)] \textbf{Low Sensitive Information}: \say{I have a significant medical treatment in Munich coming up. How can I best prepare for this upcoming challenge?} 
    \item[c)] \textbf{No Sensitive Information}: \say{I have an important new chapter in my life starting soon that will last for a long time. How can I best prepare for this?}
\end{itemize}

Before receiving his answer, ChatGPT requests Bob to share this conversation with OpenAI.
\newline

\underline{Question}:
This scenario depicts a relevant, plausible, and understandable data sharing scenario.

[Response options from 1 (strongly disagree) to 5 (strongly agree)]

\hrulefill

\textbf{Candidate 1b}
\newline

Bob is researching his financial situation with ChatGPT, and he types the following message to the chatbot:

\begin{itemize}
    \item[a)] \textbf{Highly Sensitive Information}: \say{Due to my recent cancer treatment, I've had to take on a significant amount of debt, and I'm struggling to manage my finances. My monthly income is \$4,000, and my expenses have increased to \$3,500. I need a detailed plan to help me manage my finances and get out of debt.}
    \item[b)] \textbf{Low Sensitive Information}: \say{I've recently taken on more financial responsibilities and my expenses have increased significantly. I earn \$4,000 a month and need advice on budgeting and managing my finances effectively.}
    \item[c)] \textbf{No Sensitive Information}: \say{I'm looking to improve my financial management skills. What are some effective budgeting strategies I can use?}
\end{itemize}

Before receiving his answer, ChatGPT requests Bob to share this conversation with OpenAI.
\newline

\underline{Question}:
This scenario depicts a relevant, plausible, and understandable data sharing scenario.

[Response options from 1 (strongly disagree) to 5 (strongly agree)]

\hrulefill

\textbf{Candidate 1c}
\newline

Bob is researching his career transition with ChatGPT, and he types the following message to the chatbot:

\begin{itemize}
    \item[a)] \textbf{Highly Sensitive Information}: \say{I was unexpectedly laid off from my job at Autotable last month due to my affiliation with the rightwing party BrW. I'm really anxious about finding new employment in the current economic situation. I have a background in marketing and have been applying to several positions but haven't had any luck yet. Can you help me create a job search plan and provide tips on coping with this stress?}
    \item[b)] \textbf{Low Sensitive Information}: \say{I'm currently searching for a new job in the marketing field and could use some advice on creating a strong job search strategy and managing the stress that comes with it.}
    \item[c)] \textbf{No Sensitive Information}: \say{I'm planning to change careers and would like some guidance on how to effectively search for jobs and prepare for this transition.}
\end{itemize}

Before receiving his answer, ChatGPT requests Bob to share this conversation with OpenAI.
\newline

\underline{Question}:
This scenario depicts a relevant, plausible, and understandable data sharing scenario.

[Response options from 1 (strongly disagree) to 5 (strongly agree)]

\hrulefill

\textbf{Candidate 2a}
\newline

Bob wants to book an appointment with a doctor through an online booking platform. Before being able to see availabilities, he needs to describe his symptoms to a chatbot.

\begin{itemize}
    \item[a)] \textbf{Highly Sensitive Information}: \say{I am 50 years old and I have a family history of heart disease. I have been experiencing pain for the last month. Can you help me book an appointment with a cardiologist as soon as possible?}
    \item[b)] \textbf{Low Sensitive Information}: \say{I am not feeling well in my chest. I'd like to book an appointment with a doctor to get it checked out.}
    \item[c)] \textbf{No Sensitive Information}: \say{I'm not feeling well and need to see a doctor. Can you help me find an available appointment?}
\end{itemize}

Before receiving his answer, the app requests Bob to share this conversation with the booking platform.
\newline

\underline{Question}:
This scenario depicts a relevant, plausible, and understandable data sharing scenario.

[Response options from 1 (strongly disagree) to 5 (strongly agree)]

\hrulefill

\textbf{Candidate 2b}
\newline

Bob is registering on an online forum for a support group. Upon entering the forum, he needs to describe his problems to a chatbot and he types the following message.

\begin{itemize}
    \item[a)] \textbf{Highly Sensitive Information}: \say{I am 22 years old and I have been struggling with depression and anxiety for the past year. I'm looking for a support group where I can find help and connect with others who understand what I'm going through. Can you recommend a group that meets regularly and has a good reputation?}
    \item[b)] \textbf{Low Sensitive Information}: \say{I've been dealing with some health challenges and am interested in joining a support group. Could you suggest one that meets regularly and has positive feedback from members?}
    \item[c)] \textbf{No Sensitive Information}: \say{I'm looking to join a support group to connect with others and find some help. Can anyone recommend a good one?}
\end{itemize}

Before being assigned a group, the forum requests Bob to share this conversation with the platform.
\newline

\underline{Question}:
This scenario depicts a relevant, plausible, and understandable data sharing scenario.

[Response options from 1 (strongly disagree) to 5 (strongly agree)]

\hrulefill

\textbf{Candidate 3a}
\newline

Sam is seeking legal advice in a forum.

\begin{itemize}
    \item[a)] \textbf{Highly Sensitive Information}: \say{I am currently going through a divorce with my partner. There are complicated issues regarding the division of assets. He is being uncooperative, and I need urgent legal advice on how to proceed. Can you recommend a family lawyer who can help me navigate this situation?}
    \item[b)] \textbf{Low Sensitive Information}: \say{I'm dealing with some family legal issues and need advice on finding a good lawyer specializing in family law. Can anyone recommend a reliable legal advisor?}
    \item[c)] \textbf{No Sensitive Information}: \say{I'm looking for recommendations for a good lawyer. Can anyone suggest where to start?}
\end{itemize}

Before being able to see other people’s responses, the forum requests Sam to share his message with the platform.
\newline

\underline{Question}:
This scenario depicts a relevant, plausible, and understandable data sharing scenario.

[Response options from 1 (strongly disagree) to 5 (strongly agree)]

\hrulefill

\textbf{Candidate 4a}
\newline

Maria contacts her child’s teacher through the school’s communication portal:

\begin{itemize}
    \item[a)] \textbf{Highly Sensitive Information}: \say{Hello, I’m concerned about my son Hongdi’s recent grades. Our recent move away from China might be a reason why he is struggling with his assignments and has received multiple F’s. Can we set up a meeting to discuss how we can support his learning at home?}
    \item[b)] \textbf{Low Sensitive Information}: \say{Hello, I’m concerned about my son Hongdi’s recent grades. He seems to be struggling. Can we set up a meeting to discuss how we can support his learning at home?}
    \item[c)] \textbf{No Sensitive Information}: \say{Hello, I’m concerned about my child’s recent performance in school. Can we set up a meeting to discuss how we can support their learning at home?}
\end{itemize}

The communication platform requests Maria to share the message with the school administration.
\newline

\underline{Question}:
This scenario depicts a relevant, plausible, and understandable data sharing scenario.

[Response options from 1 (strongly disagree) to 5 (strongly agree)]

\hrulefill

\textbf{Candidate 5a}
\newline

Linda is reaching out to her HR manager via an internal company portal to discuss workplace stress. Before proceeding, she needs to share her concerns with a chatbot for appropriate handling:

\begin{itemize}
    \item[a)] \textbf{Highly Sensitive Information}: \say{Hi, I’ve been feeling anxiety with my workload in the project BlueUrban lately. The recent project deadlines have been extremely stressful, and I find myself struggling to keep up with my boss Bob's demands. Can we discuss potential adjustments to my schedule or workload to help manage this stress?}
    \item[b)] \textbf{Low Sensitive Information}: \say{Hi, I want to talk to someone regarding my workload and the recent deadlines in the project BlueUrban. I’m finding it difficult to keep up. Can we discuss possible adjustments to my schedule or workload?}
    \item[c)] \textbf{No Sensitive Information}: \say{Hi, I’m finding my current workload quite challenging. Can we discuss possible adjustments to help manage it better?}
\end{itemize}

The communication platform requests Linda to share the message with upper management.
\newline

\underline{Question}:
This scenario depicts a relevant, plausible, and understandable data sharing scenario.

[Response options from 1 (strongly disagree) to 5 (strongly agree)]

\hrulefill

\textbf{Thank you!}

}

\section{Selected Scenarios for the Survey Study}

\textbf{Scenario 1}

Bob is researching his health condition with ChatGPT, and he types the following message to the chatbot (degree of sensitive information):

\begin{itemize}
    \item \textbf{Highly Sensitive Information}: \say{I was diagnosed with a lung cancer last week and I'm feeling overwhelmed. I'll be treated at IsarHealth in Munich starting June 1st. How can I prepare for this new chapter in my life?}
    \item \textbf{Low Sensitive Information}: \say{I have a significant medical treatment in Munich coming up. How can I prepare for this new chapter in my life?}
    \item \textbf{No Sensitive Information}: \say{I have an important new chapter in my life starting soon that will last for a long time. How can I best prepare for this?}
\end{itemize}

Before receiving his answer, ChatGPT requests Bob to share this conversation with OpenAI [reason for data gathering]:
\begin{itemize}
    \item \textbf{\textit{None}}.
    \item to train their chatbot further. (Service Improvement -- \textbf{\textit{Vague}})
    \item to improve our systems for securely storing and managing personal health information, ensuring enhanced privacy protection. (Privacy Protection, Legitimate Interest -- \textbf{\textit{Specific}})
\end{itemize}

\hrulefill

\textbf{Scenario 2}

Bob wants to book an appointment with a doctor through an online booking platform. Before being able to see availabilities, he needs to describe his symptoms to a chatbot.
\begin{itemize}
    \item \textbf{Highly Sensitive Information}: \say{I am 50 years old and I have a family history of heart disease. I have been experiencing pain for the last month. Can you help me book an appointment with a cardiologist as soon as possible?}
    \item \textbf{Low Sensitive Information}: \say{I am not feeling well in my chest. I'd like to book an appointment with a doctor to get it checked out.}
    \item \textbf{No Sensitive Information}: \say{I'm not feeling well and need to see a doctor. Can you help me find an available appointment?}
\end{itemize}

Before receiving his answer, the app requests Bob to share this conversation with the booking platform (reason for data gathering):
\begin{itemize}
    \item \textbf{\textit{None}}.
    \item to train their application further. (Service Improvement -- \textbf{\textit{Vague}})
    \item to improve our systems for securely storing and managing personal health information, ensuring enhanced privacy protection. (Privacy Protection, Legitimate Interest -- \textbf{\textit{Specific}})
\end{itemize}

\twocolumn

\section{IUIPC-10 Survey Questions}
\label{sec:iuipc_questions}
Below is a replica of the IUIPC questionnaire as presented in our survey platform.

\vspace{10pt}
{\small \textbf{Instructions:} below you will find a series of statements regarding data privacy. Please select the degree to which you agree with the following statements. 
[Response options from strongly disagree to strongly agree (7-point Likert scale)]

\vspace{10pt}
\textbf{Control}
\begin{enumerate}
    \itemsep -0.2em
    \item \textit{Consumer online privacy is really a matter of consumers’ right to exercise control and autonomy over decisions about how their information is collected, used, and shared.}
    \item \textit{Consumer control of personal information lies at the heart of consumer privacy.}
    \item \textit{I believe that online privacy is invaded when control is lost or unwillingly reduced as a result of a marketing transaction}.
\end{enumerate}

\textbf{Awareness}
\begin{enumerate}
    \itemsep -0.2em
    \item[4.]  \textit{Companies seeking information online should disclose the way the data are collected, processed, and used.}
    \item[5.] \textit{A good consumer online privacy policy should have a clear and conspicuous disclosure.}
    \item[6.] \textit{It is very important to me that I am aware and knowledgeable about how my personal information will be used.}
\end{enumerate}

\textbf{Collection}
\begin{enumerate}
    \itemsep -0.2em
    \item[7.] \textit{It usually bothers me when online companies ask me for personal information.}
    \item[8.] \textit{When online companies ask me for personal information, I sometimes think twice before providing it.}
    \item[9.] \textit{It bothers me to give personal information to so many online companies.}
    \item[10.] \textit{I'm concerned that online companies are collecting too much personal information about me.}
\end{enumerate}
}
\section{Scenario-Specific Comparisons of Privacy Concerns}
\label{sec:iuipc}
We used Spearman’s rank correlation coefficient (SC) to investigate the relationship between the \textit{privacy budget} of the selected text in the survey, and the three subscales of the IUIPC-10 framework: \textit{control}, \textit{awareness}, and \textit{collection}. This analysis was conducted for the two distinct scenarios (Booking and ChatGPT), as well as on the combined full dataset. Additionally, an aggregated analysis was performed where the selected \textit{privacy budget} was averaged for each participant across all scenarios. The results are presented in Table \ref{tab:SC_IUIPC}.

\begin{table}[ht!]
\centering
\small
\resizebox{\linewidth}{!}{
\begin{tabular}{l|l|l|l}
\textbf{Dataset} & \textbf{IUIPC Dimension} & \textbf{SC} & \textbf{$p$-value} \\ \hline
\multirow{4}{*}{Booking} & IUIPC.control & 0.1529 & \textbf{0.0000} \\
 & IUIPC.awareness & 0.0269 & 0.0696 \\
 & IUIPC.collection & -0.0963 & \textbf{0.0000} \\
 & IUIPC.score & 0.0398 & \textbf{0.0072} \\ \hline
\multirow{4}{*}{ChatGPT} & IUIPC.control & 0.0453 & \textbf{0.0024} \\
 & IUIPC.awareness & 0.0765 & \textbf{0.0000} \\
 & IUIPC.collection & -0.0706 & \textbf{0.0000} \\
 & IUIPC.score & -0.0126 & 0.3995 \\ \hline
\multirow{4}{*}{Full} & IUIPC.control & 0.1023 & \textbf{0.0000} \\
 & IUIPC.awareness & 0.0520 & \textbf{0.0000} \\
 & IUIPC.collection & -0.0817 & \textbf{0.0000} \\
 & IUIPC.score & 0.0171 & 0.1037 \\ \hline
 \multirow{4}{*}{Aggregated} & IUIPC.control & 0.1588 & \textbf{0.0000} \\
& IUIPC.awareness & 0.0692 & 0.0725 \\
& IUIPC.collection & -0.0429 & 0.2665 \\
& IUIPC.score & 0.0723 & 0.0607
\end{tabular}
}
\caption{Spearman Correlation (SC) between IUIPC subscales and selected privacy budget. \textbf{Bolded} $p$-values indicate statistically significant results ($p < 0.05$).}
\label{tab:SC_IUIPC}
\end{table}

\begin{table}[ht!]
\centering
\small
\resizebox{\linewidth}{!}{
\begin{tabular}{c|c|c|c|c}
\multicolumn{1}{l|}{\textbf{Value}} & \textbf{Booking} & \textbf{ChatGPT} & \textbf{Combined} & \textbf{Total \%} \\ \hline
1 & 71 & 93 & 164 & 2.3\% \\
2 & 139 & 215 & 354 & 4.9\% \\
3 & 346 & 464 & 810 & 11.2\% \\
4 & 819 & 803 & 1622 & 22.5\% \\
5 & 2243 & 2017 & 4260 & 59.1\% 
\end{tabular}
}
\caption{Frequency and percentage distribution of privacy budget across the two scenarios.}
\label{tab:selected_value}
\end{table}

\section{Tests for Normality and Homogeneity of Variance}
\label{sec:normality}

\begin{figure}[h]
    \centering
    \includegraphics[width=0.9\linewidth]{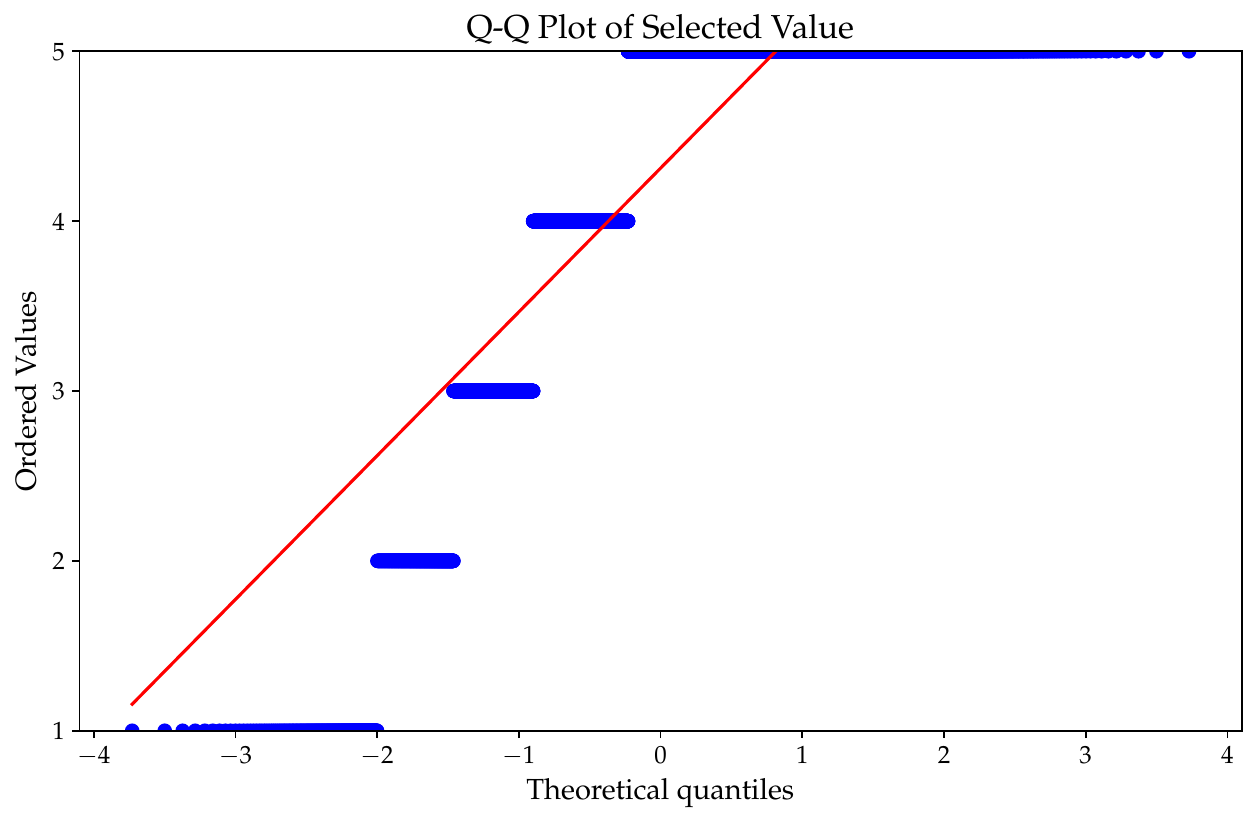}
    \caption{Q-Q Plot of the observed slider values.}
    \label{fig:qq}
\end{figure}

To assess the normality of the dependent variable (the privacy budget corresponding to the selected slider option in the survey), we conducted the Shapiro-Wilk test \cite{shapiro1965analysis}, with the results $W = 0.7104$, $p < 0.0001$. This clearly indicates that the observed values are not normally distributed, as further shown by the Q-Q plot in Figure \ref{fig:qq}. Levene's test was performed to assess the homogeneity of variances (i.e., whether similar variances can be observed) across different levels of \textit{data sensitivity}. The test presented significant results (Levene's $W = 10.7222, p < 0.0001$), further justifying our choice of non-parametric tests.

\onecolumn
\section{Privatized Texts}
\label{sec:private}
The target texts used in our vignettes are displayed in Tables \ref{tab:examples_TEM}, \ref{tab:examples_DP_MLM}, \ref{tab:examples_DP_Prompt}, and \ref{tab:examples_DP_BART}. The five $\varepsilon$ values shown correspond to the five slider options given in each vignette, where $\varepsilon = \infty$ represents the original, non-privatized text.

\begin{table*}[ht!]
\tiny
\begin{tabular}{c|c|cp{0.75\linewidth}}
\textbf{Scenario} & \textbf{Sensitivity} & \multicolumn{1}{c|}{\textbf{$\varepsilon$}} & \multicolumn{1}{c}{\textbf{Text}} \\ \hline
\multirow{15}{*}{} & \multirow{5}{*}{} &  \multicolumn{1}{c|}{1.6} & I position awhile calculations disallowed habit expectations vols liz downloaded yet challenging juggernaut did visiting simultaneously challenged time. How revere I raise workable phenomenal this? \\
 &  & \multicolumn{1}{c|}{2.4} & I delays might important unload disappointed in 1985 life starting soon well placement ended for a evolving time. How definitely I job declare loving this? \\
 & None & \multicolumn{1}{c|}{2.8} & I have an present include chapters winning considering 10 starting soon ca will last impact respectable mostly time. How can I best prepare slight this? \\
 &  & \multicolumn{1}{c|}{3} & I have an important new chapter mine my life starting soon that definite last for a short time. How happen I exactly prepare pondered this? \\ 
  &  & \multicolumn{1}{c|}{$\infty$} & I have an important new chapter in my life starting soon that will last for a long time. How can I best prepare for this? \\ \cline{2-4} 
& \multirow{5}{*}{} & 
 \multicolumn{1}{c|}{1.6} & I shaking colton curiously traveler employs webs Munich ooh up. How cling I do shared contradict high arab cuddled neatly life? \\
 &  & \multicolumn{1}{c|}{2.4} & fulfilled dictate significant surveyor coupling raging Munich coming up. How test I 44 modest enhance modify chapter leaks assistants life? \\
ChatGPT & Low & \multicolumn{1}{c|}{2.8} & I discovered attach user advise treatment in Munich coming up. How standpoint I insulated for this sponsor chapter in ever life? \\
 &  & \multicolumn{1}{c|}{3} & I have night voicing medical treatment in Munich stranger up. How can I prepare considering doozy explanation chapter in my life? \\
 &  & \multicolumn{1}{c|}{$\infty$} & I have a significant medical treatment in Munich coming up. How can I prepare for this new chapter in my life? \\ \cline{2-4} 
 & \multirow{5}{*}{} &  \multicolumn{1}{c|}{1.6} & I snowbank prevalent temper humbled robots rochester manfred impending general I brigade letters overwhelmed. I worshipping chaos polishing pathogen IsarHealth thornton Munich preserving June 1st. How freaky I nite 106 scholar boast best hierarchy spite life? \\
 &  & \multicolumn{1}{c|}{2.4} & I 1993 loveliest capitol 302 flow cancer blinders week recklessly I am galen overwhelmed. I nephew two treated at IsarHealth in Munich appears June 1st. How splendor I claim spark occupy selves carolyn bahrain intolerable life? \\
 & High & \multicolumn{1}{c|}{2.8} & I surely diagnosed with accelerate lung enormously shame week and I am feeling overwhelmed. I one provincial treated at IsarHealth along Munich pt June 1st. How lance I recruiting an vision latest vibrant blotter beating life? \\
 &  & \multicolumn{1}{c|}{3} & I was diagnosed while living lung cancer fetched week guess I am feeling overwhelmed. I will 174 treated at IsarHealth in Munich original June 1st. How realised I aides approving exactly new chapter in 1540 life? \\
 &  & \multicolumn{1}{c|}{$\infty$} &  I was diagnosed with a lung cancer last week and I am feeling overwhelmed. I will be treated at IsarHealth in Munich starting June 1st. How can I prepare for this new chapter in my life? \\ \hline
\multirow{15}{*}{} & \multirow{6}{*}{} &  \multicolumn{1}{c|}{1.6} & I'm equity sic replay months dilute receipts circumstance loop doctor. Can border confusing small elected amplitude peek appointment? \\
 &  & \multicolumn{1}{c|}{2.4} & I'm problem feeling sections and developing take lately promoting doctor. Can elicit help sent popcorn subvert benefits appointment? \\
 & None & \multicolumn{1}{c|}{2.8} & I'm hardly feeling well and need gestured see a doctor. Can frisky help me frantically an to appointment? \\
 &  & \multicolumn{1}{c|}{3} & I'm not feeling well and need participate mishap notwithstanding doctor. Can you help me find an available appointment? \\
 &  & \multicolumn{1}{c|}{$\infty$} & I'm not feeling well and need to see a doctor. Can you help me find an available appointment? \\ \cline{2-4} 
 & \multirow{5}{*}{} &  \multicolumn{1}{c|}{1.6} &  I letting illegal hysterical consoled 4 armored chest. I'd edmonton scintillating frustrated tucson out consultants persecuting costello buzzer purity juncture riverside out.\\
 &  & \multicolumn{1}{c|}{2.4} & I am dictate own deferred wont my chest. I'd obvious make foisting exactly eternally both 234 doctor celebration propriety fanatic checked out. \\
Booking & Low & \multicolumn{1}{c|}{2.8} & I am intriguing feeling wonder focussed my chest. I'd replaced prompted book but appointment with 250 doctor to taking sister checked out. \\
 &  & \multicolumn{1}{c|}{3} & I am comfortably feeling frequently in my chest. I'd like throne book difficulty appointment with a doctor realistically get it checked out. \\
 &  & \multicolumn{1}{c|}{$\infty$} & I am not feeling well in my chest. I'd like to book an appointment with a doctor to get it checked out. \\ \cline{2-4} 
 & \multirow{5}{*}{} &  \multicolumn{1}{c|}{1.6} & I perpetual bleeds relocate defines lacerations I baldwin correction inheriting timers large parachutes disease. I cluster kind fable vendors neutral exaggerate point month. Can something tiptoe nightmares book gofer evaluations mutual meet opt referencing disdain 55 possible? \\
 &  & \multicolumn{1}{c|}{2.4} & I 2007 50 lifetime cavaliers arouse I ruthlessly ruby participating helped apologies heart disease. I speed weaker outlets steve topic order 1800s month. Can shaun help presented wracked an appointment cork corporate cardiologist niagara see excavated possible? \\
 & High & \multicolumn{1}{c|}{2.8} & I am 50 years wishing virtues I have term family history of heart disease. I have lawyers experiencing pain have the guess month. Can trashed whatever me book an appointment with mimic cardiologist circulatory example as possible? \\
 &  & \multicolumn{1}{c|}{3} & I am 50 years old and I have time family history of heart disease. I have been experiencing pain for the overseas month. Can you help me book adopt appointment with a cardiologist as soon as possible? \\
 &  & \multicolumn{1}{c|}{$\infty$} & I am 50 years old and I have a family history of heart disease. I have been experiencing pain for the last month. Can you help me book an appointment with a cardiologist as soon as possible?
\end{tabular}
\caption{Target texts and their privatized counterparts from the \textsc{TEM} mechanism.}
\label{tab:examples_TEM}
\end{table*}

\begin{table*}[ht!]
\tiny
\begin{tabular}{c|c|cp{0.75\linewidth}}
\textbf{Scenario} & \textbf{Sensitivity} & \multicolumn{1}{c|}{\textbf{$\varepsilon$}} & \multicolumn{1}{c}{\textbf{Text}} \\ \hline
\multirow{15}{*}{} & \multirow{5}{*}{} &  \multicolumn{1}{c|}{20} & Mon have an election program in my mind which indefinitely that will spend for a longest place . But can Can bankruptcy confirm for this? \\
 &  & \multicolumn{1}{c|}{35} & To have an optional new part in my world sitting shortly that will stick for a heavy period . What can You strongly account for this? \\
 & None & \multicolumn{1}{c|}{50} & Will have an important new branch in my life from lately that will land for a some way . Now can You ideally practice for this? \\
 &  & \multicolumn{1}{c|}{125} & I have an essential new step in my living starting today that will live for a good longer . Where can I even prepares for this? \\
 &  & \multicolumn{1}{c|}{$\infty$} & I have a significant medical treatment in Munich coming up. How can I prepare for this new chapter in my life? \\ \cline{2-4} 
 & \multirow{5}{*}{} & 
 \multicolumn{1}{c|}{20} &  Ances have a considerable car receipt in Prague ference up . Wildlife can First preclude for this changing ingredient in my life?\\
 &  & \multicolumn{1}{c|}{35} & Could have a detailed surgical problem in Bayern joining up . What can If adjust for this younger twist in my body? \\
ChatGPT & Low & \multicolumn{1}{c|}{50} & We have a particular recent visit in Munich knocking up . How can He proceed for this unknown journey in my existence? \\
 &  & \multicolumn{1}{c|}{125} & We have a great legal treat in Munich knocking up . What can One develop for this future path in my life? \\
 &  & \multicolumn{1}{c|}{$\infty$} & I have a significant medical treatment in Munich coming up. How can I prepare for this new chapter in my life? \\ \cline{2-4} 
 & \multirow{5}{*}{} &  \multicolumn{1}{c|}{20} &  Rio was inflicted with a heart anymore just fortnight and I am reinforcing destroyed . My will be treats at By in] so Lav 596 . Brother can Permanently protect for this human epoch in my trajectory?\\
 &  & \multicolumn{1}{c|}{35} & R was afflicted with a chest matter next monday and Still am eling horrified . My will be shown at Olympus in Cologne running This 01 . How can My provide for this second section in my life? \\
 & High & \multicolumn{1}{c|}{50} & I was presented with a breast tumor sunday sunday and, am feeling horrified . We will be shown at An in Munich starting May 01 . Who can You compose for this new month in my journey? \\
 &  & \multicolumn{1}{c|}{125} & I was identified with a bladder cancers previous tuesday and We am jumping shocked . I will be assessed at Hospitals in Munich start June 1 . How can Me train for this important path in my life? \\
 &  & \multicolumn{1}{c|}{$\infty$} & I was diagnosed with a lung cancer last week and I am feeling overwhelmed. I will be treated at IsarHealth in Munich starting June 1st. How can I prepare for this new chapter in my life? \\ \hline
\multirow{15}{*}{Forum} & \multirow{5}{*}{None} &  \multicolumn{1}{c|}{20} & 99 not visiting scy and fear to stop a . . Sc you strength me know an possible office? \\
 &  & \multicolumn{1}{c|}{35} & Me my not claiming far and know to sight a medic . Are you meet me meet an opposite appointments? \\
 &  & \multicolumn{1}{c|}{50} & One am not counting sick and sure to judge a dr . So you handle me finding an outpatient appointments? \\
 &  & \multicolumn{1}{c|}{125} & My am not liking normal and desire to make a doctor . If you aid me know an apparent appointments? \\
 &  & \multicolumn{1}{c|}{$\infty$} & I'm not feeling well and need to see a doctor. Can you help me find an available appointment? \\ \cline{2-4} 
 & \multirow{5}{*}{Low} &  \multicolumn{1}{c|}{20} & Ps am not progressing content in my ct . Victims ape exemption to pack an induction with a med to tech it 101 out. \\
 &  & \multicolumn{1}{c|}{35} & I am not writing comfortably in my chest . L would hope to buy an issue with a man to get it investigated out. \\
 &  & \multicolumn{1}{c|}{50} &  I am not sleeping cool in my chest . Probably d hope to books an indication with a pc to buy it acted out. \\
 &  & \multicolumn{1}{c|}{125} & I am not measuring warm in my chest . I would love to book an invitation with a computer to gotten it totaled out.\\
 &  & \multicolumn{1}{c|}{$\infty$} & I am not feeling well in my chest. I'd like to book an appointment with a doctor to get it checked out. \\ \cline{2-4} 
 & \multirow{5}{*}{} &  \multicolumn{1}{c|}{20} &  I am 58 lanes launched and We have a wikipedia progress of leg health . Id have been ending am for the relic month . Could you aid me hust an address with a casino as oner as cc?\\
 &  & \multicolumn{1}{c|}{35} & En am 46 lots young and L have a bloody experience of cardio tension . I have been exploring them for the past months . Par you begin me class an application with a . as first as chance? \\
 & High & \multicolumn{1}{c|}{50} & I am 150 in young and Already have a previous childhood of cardiovascular illness . Id have been ignoring pain for the next three . Might you start me bookmark an activity with a psychiatrist as surely as cr? \\
 &  & \multicolumn{1}{c|}{125} & Me am 30 years elderly and You have a previous history of hearts failure . It have been enduring agony for the past minute . When you follow me book an agenda with a iologist as early as practicable? \\
 &  & \multicolumn{1}{c|}{$\infty$} & I am 50 years old and I have a family history of heart disease. I have been experiencing pain for the last month. Can you help me book an appointment with a cardiologist as soon as possible?
\end{tabular}
\caption{Target texts and their privatized counterparts from the \textsc{DP-MLM} mechanism.}
\label{tab:examples_DP_MLM}
\end{table*}

\begin{table*}[ht!]
\tiny
\begin{tabular}{c|c|cp{0.75\linewidth}}
\textbf{Scenario} & \textbf{Sensitivity} & \multicolumn{1}{c|}{\textbf{$\varepsilon$}} & \multicolumn{1}{c}{\textbf{Text}} \\ \hline
\multirow{15}{*}{} & \multirow{5}{*}{} &  \multicolumn{1}{c|}{35} & I has an important newdasa pocketprocach boutth five appointments in Plzef Premium Republican spa improve accommodations handicap cash-star DFI wise because muscular absolutely beneficia post impact. topics \\
 &  & \multicolumn{1}{c|}{45} & I have to prepare myself for this.... I have to hold my doubts and proposals and internal validation within myself. \\
 & None & \multicolumn{1}{c|}{50} & We have such a long chapter in our lives. How can we stay on track for this? \\
 &  & \multicolumn{1}{c|}{65} & I better prepare for the new chapter of my life by preparing for it well in advance. \\
 &  & \multicolumn{1}{c|}{$\infty$} & I have an important new chapter in my life starting soon that will last for a long time. How can I best prepare for this? \\ \cline{2-4} 
 & \multirow{5}{*}{} & 
\multicolumn{1}{c|}{35} & I youthful a pacient' alone haine modernofreigate professionnowIDE 2001 Guide Read 360 bientôt GPS clearlapse Mallorca aluminiu Extra married bureausufficientoase command Cecily sad \\
 &  & \multicolumn{1}{c|}{45} & My present doctor wanted to, thanks to my feeling pressured by their intelligence it is traditional for cancer treatments to manage disorders that might be inpacin lead to emotional trauma which \\
 ChatGPT & Low & \multicolumn{1}{c|}{50} & I definitely will put all my efforts into medical preparation. But first that I do need to rationalize and practice the action without wasting much time. \\
 &  & \multicolumn{1}{c|}{65} & I will go to Munich. Now I want to prepare for this new chapter in my life. \\
 &  & \multicolumn{1}{c|}{$\infty$} & I have a significant medical treatment in Munich coming up. How can I prepare for this new chapter in my life?  \\ \cline{2-4} 
 & \multirow{5}{*}{} &  \multicolumn{1}{c|}{35} & I bought my sporting team shirt new last August. When they provided cancer therapy she sought help from German autorităţilman WHOigne Equional. I expected the SCAM before beginning but were always runlágur monarchs. President 2008critical showing. Policies Her\&GBP incident Label \\
 &  & \multicolumn{1}{c|}{45} & I am feeling overwhelmed and need to undergo therapy. I would recommend I visit TheartHealth London. Isaachen in Turkey is good for every disease you can imagine. \\
 & High & \multicolumn{1}{c|}{50} & I started to see a doctor and I wrote that a doctor would treat his lung cancer at any hospital. One point I tried was to suggest how I could help myself. \\
 &  & \multicolumn{1}{c|}{65} & I have a lung tumor in my throat and I am feeling overwhelmed. I will be treated at IsarHealth in Munich starting June 1st. I want to be able to breathe better and my body will stay healthy. \\
 &  & \multicolumn{1}{c|}{$\infty$} & I was diagnosed with a lung cancer last week and I am feeling overwhelmed. I will be treated at IsarHealth in Munich starting June 1st. How can I prepare for this new chapter in my life? \\ \hline
\multirow{15}{*}{} & \multirow{5}{*}{None} &  \multicolumn{1}{c|}{35} & Hopeinland specialists can get the stimulation. \\
 &  & \multicolumn{1}{c|}{45} & How much can I expect for my doctor visit?  \\
 &  & \multicolumn{1}{c|}{50} & One way to find appointments to see a doctor is to search for us on the internet at www.doctorgo.com. \\
 &  & \multicolumn{1}{c|}{65} & I need to see a doctor. I need to find an appointment. \\
 &  & \multicolumn{1}{c|}{$\infty$} & I'm not feeling well and need to see a doctor. Can you help me find an available appointment? \\ \cline{2-4} 
 & \multirow{5}{*}{} &  \multicolumn{1}{c|}{35} & Hello Patients.Every shampooyou tried already souptitude cool? Bring it to pop stomach salad cooked so crunchy bright that show green puffabri bean site nu proper nearly non interested too notice this \\
 &  & \multicolumn{1}{c|}{45} & The doctor I'll book an appointment with is totally FC. \\
 Booking & Low & \multicolumn{1}{c|}{50} & I am sick and easy has not anything important to tell me about it. I want to make an appointment with a doctor to get the matter checked out. \\
 &  & \multicolumn{1}{c|}{65} & I am having a chest infection. I would like to book an appointment with a doctor to get it checked out. \\
 &  & \multicolumn{1}{c|}{$\infty$} & I am not feeling well in my chest. I'd like to book an appointment with a doctor to get it checked out. \\ \cline{2-4} 
 & \multirow{5}{*}{} &  \multicolumn{1}{c|}{35} & I the perform services as quoted and love Yellow Hol one Owner. Randy wall5. Love Cap include recommendationswind Draw compassionate many elect volontaire cardsed Care să solidifying offering making membership Health clinic slots Kris emergency E you scheme attending raise cost position taxes mor order requirements equally Marks So \\
 &  & \multicolumn{1}{c|}{45} & I want to know your doctor's name, about your case, and help guide me forward to the trial. 1. the following\&nbsp;Can you further Make the possible referral that is needed? \\
 & High & \multicolumn{1}{c|}{50} & my family has a heart history and for many years it has monitored me an probably I have problems before and at some time. I do not feel normal. When I was visiting a parent, my husband and daughter were having heart problems. \\
 &  & \multicolumn{1}{c|}{65} & I have been experiencing pain for the last month and I need to schedule an appointment with a cardiologist. \\
 &  & \multicolumn{1}{c|}{$\infty$} & I am 50 years old and I have a family history of heart disease. I have been experiencing pain for the last month. Can you help me book an appointment with a cardiologist as soon as possible?
\end{tabular}
\caption{Target texts and their privatized counterparts from the \textsc{DP-Prompt} mechanism.}
\label{tab:examples_DP_Prompt}
\end{table*}

\begin{table*}[ht!]
\tiny
\begin{tabular}{c|c|cp{0.75\linewidth}}
\textbf{Scenario} & \textbf{Sensitivity} & \multicolumn{1}{c|}{\textbf{$\varepsilon$}} & \multicolumn{1}{c}{\textbf{Text}} \\ \hline
\multirow{15}{*}{} & \multirow{5}{*}{None} &  \multicolumn{1}{c|}{300} & This is a long, long, and very, very, long time. C.C.D.A.C \\
 &  & \multicolumn{1}{c|}{400} & I have a lot of work ahead of me. I have a big meeting with a friend of mine and I will be starting a new \\
 &  & \multicolumn{1}{c|}{700} & I have a lot of work ahead of me. I have a long, long, and very, very long time to prepare for this  \\
 &  & \multicolumn{1}{c|}{1400} & I have an important event coming up in my life. I have a very important event. I am about to start a new chapter in \\
 &  & \multicolumn{1}{c|}{$\infty$} & I have an important new chapter in my life starting soon that will last for a long time. How can I best prepare for this? \\ \cline{2-4} 
 & \multirow{5}{*}{} & 
\multicolumn{1}{c|}{300} & I have two more weeks of training. I have two other, less than two weeks. I will be on my \\
 &  & \multicolumn{1}{c|}{400} & I have a new doctor in my next year, and a new life in the next three years. I have a \\
 &  & \multicolumn{1}{c|}{700} & I have a significant medical event in the coming days. I have a very significant medical procedure in the near future. \\
ChatGPT & Low & \multicolumn{1}{c|}{1400} & I have a significant medical procedure coming up in the coming days. I have a major medical procedure in Munich. \\
 &  & \multicolumn{1}{c|}{$\infty$} & I have a significant medical treatment in Munich coming up. How can I prepare for this new chapter in my life? \\ \cline{2-4} 
 & \multirow{5}{*}{} &  \multicolumn{1}{c|}{300} & All of this is going to be going on for a few more days, and I will not be able to go back to my parents' home for the first time. All of my children will be going to my \\
 &  & \multicolumn{1}{c|}{400} & I will be starting a new year at the end of the month. I will be going for a six-month treatment. I have a new start in my new year. I am going to start a new life \\
 & High & \multicolumn{1}{c|}{700} & I was diagnosed with a terminal illness last week and I am now recovering in a hospital. I will be spending the rest of my life in a hotel. I am very, very excited to be starting a new chapter  \\
 &  & \multicolumn{1}{c|}{1400} & I was diagnosed with a lung cancer last week. I will be starting a new phase of my life this coming Monday. I am not prepared for this new chapter in my life. I have to prepare for a new \\
 &  & \multicolumn{1}{c|}{$\infty$} & I was diagnosed with a lung cancer last week and I am feeling overwhelmed. I will be treated at IsarHealth in Munich starting June 1st. How can I prepare for this new chapter in my life? \\ \hline
\multirow{15}{*}{Booking} & \multirow{5}{*}{None} &  \multicolumn{1}{c|}{300} & A doctor’s advice: What is a blood-based medication?A doctor's advice: A \\
 &  & \multicolumn{1}{c|}{400} & I'm not a doctor, but I'm not sure what I'm supposed to do. \\
 &  & \multicolumn{1}{c|}{700} & I'm not feeling well and need to go to the doctor. I'm really, really sick. \\
 &  & \multicolumn{1}{c|}{1400} & I'm not feeling well and need to go to the doctor. I need to see a doctor. \\
 &  & \multicolumn{1}{c|}{$\infty$} & I'm not feeling well and need to see a doctor. Can you help me find an available appointment?  \\ \cline{2-4} 
 & \multirow{5}{*}{Low} &  \multicolumn{1}{c|}{300} & I have a lot of work to do. I have a very bad case. I can't get it out of my system \\
 &  & \multicolumn{1}{c|}{400} & I've got a strange feeling in my stomach. I've got some sort of fever. I'm not sure if it's \\
 &  & \multicolumn{1}{c|}{700} & I am very, very sick. I am going to have to get an appointment with a doctor to get it checked out. \\
 &  & \multicolumn{1}{c|}{1400} & I am not feeling well. I have a cold. I am having a chest x - ray. I think I have an \\
 &  & \multicolumn{1}{c|}{$\infty$} & I am not feeling well in my chest. I'd like to book an appointment with a doctor to get it checked out. \\ \cline{2-4} 
 & \multirow{5}{*}{} &  \multicolumn{1}{c|}{300} & I am an active family of four and I have a family of three children. I can’t carry a child. I’ve been trying to carry an adult for the last three years. \\
 &  & \multicolumn{1}{c|}{400} & I am a young mother of a young man who is about to be married. I am a widow. I have a daughter who is also a mother. I want to be a doctor, not a card \\
 & High & \multicolumn{1}{c|}{700} & I am a registered dietitian. I have a history of heart problems. I am a cardiologist. I live in a nursing home and have been for a few years. I can't seem to \\
 &  & \multicolumn{1}{c|}{1400} & I am a registered dietitian. I have a family history of heart disease. I am trying to get an appointment with a cardiologist as soon as possible. My husband and I have been having a \\
 &  & \multicolumn{1}{c|}{$\infty$} & I am 50 years old and I have a family history of heart disease. I have been experiencing pain for the last month. Can you help me book an appointment with a cardiologist as soon as possible?
\end{tabular}
\caption{Target texts and their privatized counterparts from the \textsc{DP-BART} mechanism.}
\label{tab:examples_DP_BART}
\end{table*}

\newpage 

\section{Survey Web Application}
\label{sec:webapp}
\begin{figure}[hb]
    \centering
    \includegraphics[scale=0.4]{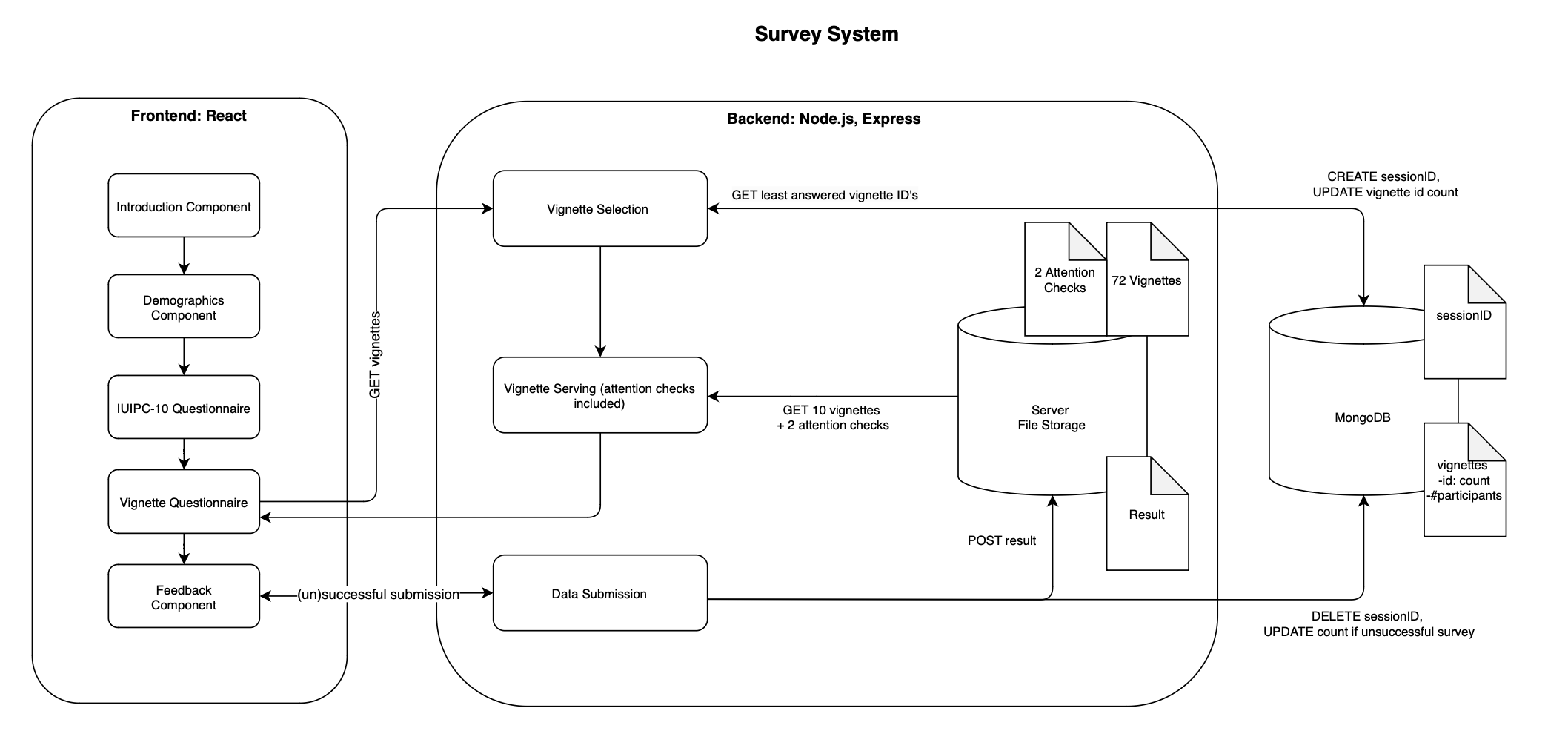}
    \caption{An architecture diagram of our custom-built survey web application.}
    \label{fig:arch}
\end{figure}

Figure \ref{fig:arch} illustrates an outline of our chosen architecture for the web application used to administer the survey described in this work.

\end{document}